\documentclass[acmtocl,acmnow]{acmtrans2m}

\usepackage{graphicx}
\usepackage{amsmath}
\usepackage{amssymb}
\usepackage{stmaryrd}

\let\emptyset=\varnothing
\let\leq=\leqslant
\let\geq=\geqslant

\newtheorem{theorem}{Theorem}[section]
\newtheorem{proposition}[theorem]{Proposition}
\newtheorem{lemma}[theorem]{Lemma}
\newdef{definition}[theorem]{Definition}
\newdef{example}[theorem]{Example}
\newdef{remark}[theorem]{Remark}

\newcommand{\itv}[2]{\ensuremath{[#1, #2]}}

\newcommand{\kwd}[1]{\textbf{#1}}
\newcommand{\algo}[1]{{\normalfont\textsf{#1}}}

\newcommand{\SWITCH}[1]{\kwd{switch} #1 \kwd{in}}
\newcommand{\RETURN}[1]{\kwd{return} #1}
\newcommand{\IF}[1]{\kwd{if} #1~\kwd{then}}
\newcommand{\ELSE}{\kwd{else}}
\newcommand{\ENDIF}{\kwd{endif}}
\newcommand{\ENDSWITCH}{\kwd{end}}
\newcommand{\BEGIN}{\kwd{begin}}
\newcommand{\END}{\kwd{end}}

\newcommand{\FOREACH}[1]{\kwd{foreach} #1~\kwd{do}}
\newcommand{\ENDFOREACH}{\kwd{end}}

\newcommand{\NUM}[1]{{\tiny{\hbox to 1em{\hfill#1}}}}

\newcommand{\GSet}[1]{\ensuremath{\mathcal{#1}}}

\newcommand{\Vector}[1]{\ensuremath{\mathbf{#1}}}
\newcommand{\SVector}[1]{\ensuremath{\boldsymbol{#1}}}

\newcommand{\cplusplus}{{C\nolinebreak[4]\hspace{-.15em}\raisebox{.47ex}{\tiny\bf ++}}}

\newcommand{\GNSet}[1]{\ensuremath{\mathbb{#1}}}
\newcommand{\NSet}{\GNSet{N}}
\newcommand{\RSet}{\GNSet{R}}
\newcommand{\FSet}{\GNSet{F}}
\newcommand{\ISet}{\GNSet{I}}

\newcommand{\powerset}[1]{\ensuremath{\GSet{P}(#1)}}

\newcommand{\prevFloat}[1]{\ensuremath{{#1}^-}}
\newcommand{\nextFloat}[1]{\ensuremath{{#1}^+}}

\newcommand{\LB}[1]{\ensuremath{\underline{#1}}}
\newcommand{\RB}[1]{\ensuremath{\overline{#1}}}

\newcommand{\replaceDom}[3]{\ensuremath{#1_{{#2}\mathbin{\leftarrow}{#3}}}}

\newcommand{\negation}[1]{\ensuremath{\overline{#1}}}

\newcommand{\BComplement}[2]{\ensuremath{#1\boxbslash #2}}

\newcommand{\Domain}[2]{\ensuremath{\algo{Dom}_{#2}(#1)}}

\newcommand{\object}[1]{\ensuremath{\mathfrak{#1}}}

\newcommand{\benchmark}[1]{\subsubsection*{\textbf{#1}}}

\newcommand{\dist}[2]{\ensuremath{\algo{dist}\bigl(#1,#2\bigr)}}

\newcommand{\Outer}[1]{\ensuremath{\algo{Outer}(#1)}}
\newcommand{\Inner}[1]{\ensuremath{\algo{Inner}(#1)}}

\newcommand{\CRel}[1]{\ensuremath{\rho_{#1}}}

\newcommand{\BoxesToSet}[1]{\ensuremath{\Lbag{#1}\Rbag}}

\title{Interval Constraint Solving\\for Camera Control and Motion Planning}

\author{%
  FR\'ED\'ERIC BENHAMOU, FR\'ED\'ERIC GOUALARD, \\ 
  \'ERIC LANGU\'ENOU, and MARC CHRISTIE \\
  Institut de Recherche en Informatique de Nantes, France}

\begin{abstract}
  Many problems in robust control and motion planning can be reduced to
  either find a sound approximation of the solution space determined by
  a set of nonlinear inequalities, or to the ``guaranteed tuning
  problem'' as defined by Jaulin and Walter, which amounts to finding
  a value for some tuning parameter such that a set of inequalities be
  verified for all the possible values of some perturbation vector. A
  classical approach to solve these problems, which satisfies the
  strong soundness requirement, involves some quantifier elimination
  procedure such as Collins' Cylindrical Algebraic Decomposition
  symbolic method. Sound numerical methods using interval arithmetic
  and local consistency enforcement to prune the search space are
  presented in this paper as much faster alternatives for both soundly solving
  systems of nonlinear inequalities, and addressing the guaranteed
  tuning problem whenever the perturbation vector has dimension one.
  The use of these methods in camera control is investigated, and
  experiments with the prototype of a declarative modeller to express
  camera motion using a cinematic language are reported and commented.
\end{abstract}


\begin{document}

\begin{bottomstuff}
  The research exposed here was supported in part by the INRIA LOCO
  project and a project of the French/Russian A.M.\ Liapunov
  Institute.
  
  This is a revised and extended version of a paper presented at the
  sixth International Conference on Principles and Practice of
  Constraint Programming (CP~'00), Singapore.
\end{bottomstuff}

\maketitle

\section{Introduction}
\label{sec:introduction}

Designing electronic circuits~\cite{Ebers-Moll:IEE54}, identifying the
structure of complex molecules~\cite{Emiris-Mourrain:Algorithmica99}, or
computing the quantity of chemical elements produced by some
reaction~\cite{Meintjes-Morgan:TOMS90} are all problems---among many
others---that can be modelled by sets of real nonlinear equations and
inequations.  \emph{Reliably} solving these systems (that is, delivering
approximate solutions as close as possible to the true ones) with the
limited set of real numbers representable on computers has generated a
large amount of literature since the very dawn of computer
science~\cite{Turing:QJM48,Rademacher:ACLHU48,Wilkinson:63}.

Following Sunaga's and Moore's seminal
works~\cite{Sunaga:58,Moore:66}, \emph{interval analysis} has been
identified as a key tool for ensuring reliability and
\emph{completeness} (that is, the capability to retain all the
solutions) in the solving process. 

Interval constraint
solving~\cite{Older-Vellino:CCECE90,Benhamou:Chatillon94} uses
interval arithmetic in algorithms alternating \emph{propagation
  steps}~\cite{Waltz:75,Mackworth:AI77}, which enforce some local
consistency notion to tighten the variables' feasible domain, and
\emph{search steps} to isolate different solutions and overcome
the weakness of the propagation steps.

Interval constraint solvers such as
clp(BNR)~\cite{Benhamou-Older:JLP97}, ILOG
Solver~\cite{Puget:SPICIS94}, or
Numerica~\cite{Van-Hentenryck-et-al:97} have been shown to be
efficient tools for solving some challenging nonlinear constraint
systems~\cite{Puget-Van-Hentenryck:JGO98,Granvilliers-Benhamou:JGO01}).
Relying on interval arithmetic, they guarantee completeness and
isolate punctual solutions with an ``arbitrary'' accuracy. They take as
input a constraint system and a Cartesian product of domains for the
variables occurring in the constraints; their output is a set
$\GSet{S}_o$ of boxes approximating each solution contained in the
input box.

However, \emph{soundness} (i.e., the property that all boxes returned
contain nothing but true solution points) is not guaranteed while it
is sometimes a strong requirement. Consider, for instance, a civil
engineering problem~\cite{Sam-Haroud:PhD95} such as floor design where
retaining non-solution points may lead to a physically infeasible
structure.  As pointed out by Ward et al.~\cite{Ward-et-al:IJCAI89}
and Shary~\cite{Shary:MISC99}, one may expect different properties
from the boxes composing $\GSet{S}_o$ depending on the problem at
hand, namely: every element in any box is a solution, or there exists
at least one solution in each box. Interval constraint solvers ensure
only, at best, the second property.

Interval constraint solving techniques also suffer from a severe limitation 
in that they can only handle constraint systems with discrete solutions.
Yet, many applications in robust control or error-bounded 
estimation~\cite{Jaulin-Walter:Automatica93} are modelled by systems of
nonlinear inequalities, whose solution set is usually not discrete.

In addition, even more applications can be reduced to what Jaulin and
Walter~\cite{Jaulin-Walter:Automatica96} call the ``guaranteed tuning
problem'', which amounts to finding the values for some tuning
parameter such that a set of inequalities be verified for all the
possible values of some perturbation vector\footnote{Jaulin and Walter
  restrict the problem to finding only one value for the tuning
  parameter. The generalization adopted here is of interest to any
  application in which the user has to be given the choice of the solution
  to adopt, such as in the camera control application presented in
  this paper.}. More precisely:
\begin{quote}\itshape
  Given \Vector{B}, a box of feasible values for some tuning parameter vector \SVector{\gamma}
  and \Vector{D}, a box of feasible values for some perturbation vector \SVector{\pi},
  find the set $\GSet{S}_{\SVector{\gamma}}$ defined by:
  \begin{equation*}
    \GSet{S}_{\SVector{\gamma}}=\{\SVector{\gamma}\in\Vector{B}\mid
                     \forall\SVector{\pi}\in\Vector{D}\colon
                           \Vector{f}(\SVector{\gamma},\SVector{\pi})\geq\Vector{0}\}
  \end{equation*}
  where \Vector{f} is a vector of nonlinear functions, and the inequality is to 
  be taken componentwise.
\end{quote}

\begin{figure}
\begin{center}
\includegraphics[width=.5\textwidth]{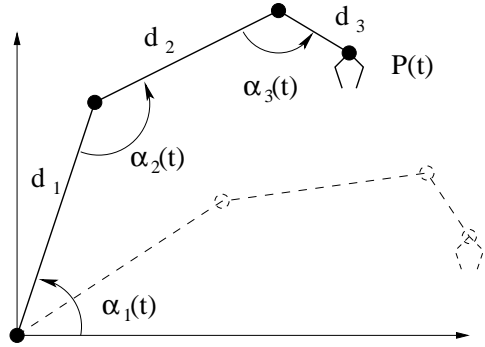}
\caption{Avoiding collisions with the arm of a robot}
\label{fig:robot}
\end{center}
\end{figure}

Problems ranging from robust control~\cite{Abdallah-et-al:MSCA96} and camera
control~\cite{Drucker-Zeltzer:GI94} to motion planning~\cite{Tarabanis:TR90} can all be formulated as
guaranteed tuning problems. Consider for example the following application:

\begin{example}[A collision-free problem]\label{exa:robot}
  A mobile robot arm is composed of three segments of respective
  lengths $d_1$, $d_2$ and $d_3$, and three motor-controlled axes (see
  Figure~\ref{fig:robot}). The trajectory of the robot's hand $P(t)$ is
  therefore determined by three angle functions $\alpha_1(t)$,
  $\alpha_2(t)$ and $\alpha_3(t)$. The problem is to find all points
  in the $2D$ space that do not collide with the robot's hand, i.e.\
  computing all the $(x,y)$ coordinates such that the distance
  between $P(t)$, for all $t$, and $(x,y)$ is greater than some given
  value $d$ (here, the size of the robot's hand):
\begin{equation*}
\forall t\in \itv{0}{1}\colon \sqrt{(x - P_x(t))^2 + (y - P_y(t))^2} \geq d
\end{equation*}
where $P_x(t)$ and $P_y(t)$ represent the coordinates of $P(t)$ at time $t$, defined by:
\begin{equation*}
\left\{\begin{array}{ll}
P_x(t) =& d_1\sin
\alpha_1(t)+d_2\sin\bigl(\alpha_1(t)+\alpha_2(t)-\pi\bigr)+
     d_3\sin\bigl(\alpha_1(t)+\alpha_2(t)+\alpha_3(t)\bigr)\\
P_y(t) =& d_1\cos\alpha_1(t)+d_2\cos\bigl(\alpha_1(t)+\alpha_2(t)-\pi\bigr)+
     d_3\cos\bigl(\alpha_1(t)+\alpha_2(t)+\alpha_3(t)\bigr)\\
\end{array}\right.
\end{equation*}
\end{example}

Until recently, the soundness issue and the presence of quantifiers
called for symbolic methods and quantifier elimination procedures such
as Cylindrical Algebraic Decomposition~\cite{Collins:CATFL75} (CAD).
Unfortunately, these techniques are either too slow, or limited to
polynomial constraints.

The advent of interval analysis led to the devising of simple and
sound algorithms to solve systems of
inequalities~\cite{Jaulin-Walter:Automatica93,Garloff-Graf:99} and the
guaranteed tuning problem (restricted to the finding of only one
value, though)~\cite{Jaulin-Walter:Automatica96}. By and large, many of
these algorithms are but sophisticated interval extensions of simple
search procedures, meaning that are they computationally expensive, even
for small problems.

In this paper, we present sound algorithms that draw upon efficient
complete interval constraint solving pruning methods to soundly solve systems
of inequalities and the guaranteed tuning problem for the case of
unidimensional perturbation vectors. This kind of problem is the one
occurring in motion planning and camera control applications, where
the only universally quantified variable is usually the time.

The outline of the paper is as follows: in order to be reasonably
self-content, the basics of interval analysis are presented in
Section~\ref{sec:interval-analysis}; related works on both solving
systems of nonlinear inequalities and the guaranteed tuning problem
are presented in Section~\ref{sec:related-work}; their strong
points and weaknesses regarding the applications targeted are also pointed
out; interval constraint solving is introduced in
Section~\ref{sec:interval-constraint-solving} as a basis for the new
sound algorithms presented in
Section~\ref{sec:sound-constraint-solving}; the modelling of a camera
control problem in terms of these new sound interval constraint
methods is then described in
Section~\ref{sec:virtual-cameraman-problem}; the results with a
prototype of a declarative modeller allowing a non-technician user to
control the positioning of a camera by means of a cinematic language
are commented in Section~\ref{sec:benchmarking} and contrasted with
the ones obtained by Jardillier and
Langu\'enou~\cite{Jardillier-Languenou:EG98} on the same problems with
a different approach; finally, Section~\ref{sec:conclusion} discusses
directions for future researches.

\section{Interval Analysis}
\label{sec:interval-analysis}

In this section, we limit the presentation of interval analysis to the
concepts needed in the sequel of the paper.  The reader is referred to
works by Moore, Hansen and
others~\cite{Moore:66,Alefeld-Herzberger:83,Hansen:92,Neumaier:90} for
a more complete presentation.

\medskip The finite nature of computers implies that they can only
represent and manipulate a small subset of the real numbers
represented in a floating-point format. Since the mid-eighties, most
computers comply with the IEEE~754 standard~\cite{IEEE754:90}
specifying the format of floating-point numbers.

The set of floating-point numbers \FSet\ is but a very small subset of
real numbers. In addition, it is not closed for arithmetic operations,
meaning that rounding of the results to representable floating-point
numbers must usually take place. We say that an operation is
\emph{correctly rounded} when the value chosen, whenever the true
result is not representable, is the closest floating-point number.
Note that the IEEE~754 standard only requires the operators
$+,-,\times,\div,\sqrt{\:}$ to be correctly rounded. The precision of
the other operators is implementation dependent. See the paper by
Lef\`evre et al.~\cite{Lefevre-et-al:ToC98} for more information on
this topic.

Given a floating-point number $a$, let \nextFloat{a} (resp.\ 
\prevFloat{a}) be the smallest float greater than $a$ (resp.\ greatest
float smaller than $a$).

Rounding makes the reliable solving of systems of nonlinear equalities
or inequalities a challenging task, to which a large amount of papers
and books has been devoted.

One solution advocated by Moore~\cite{Moore:66} to control the rounding errors is
to use \emph{interval arithmetic}, that is to replace reals by intervals
containing them, whose bounds are representable numbers. For example:
\begin{equation*}
  \pi\in\itv{3.14}{3.15}
\end{equation*}

An interval $I$ with representable bounds is called a
\emph{floating-point interval}. It is of the form:
$I=\itv{a}{b}=\{r\in\RSet\mid a\leq r\leq b,\:\text{with}\:\:
a,b\in\FSet\}$.  A non-empty interval $I=\itv{a}{b}$ such that $b\leq
\nextFloat{a}$ is said \emph{canonical}. In the same way, a Cartesian
product of intervals (or \emph{box}) is said canonical whenever it is
canonical in all its dimensions. In the sequel, vectors or Cartesian products
are written in bold face.

Let \ISet\ be the set of all floating-point
intervals. Since we will only deal with this kind of intervals in the
rest of the paper, we will refer to them as simply \emph{intervals}.

Operations and functions over reals are also replaced by an
\emph{interval extension} having the \emph{containment property}.
Given a real function, $f\colon\RSet^n\to\RSet$ and a box
$\Vector{B}\in\ISet^n$, let $f(\Vector{B})=\{f(\Vector{r})\mid
\Vector{r}\in\Vector{B}\}$, $\mathcal{D}_f$ be the domain of $f$, and
$\mathcal{D}^{\ISet}_f=\{\Vector{B}\in\ISet^n\mid
\Vector{B}\subseteq\mathcal{D}_f\}$.

\begin{definition}[Interval extension]
  Given a real function $f\colon \RSet^n\to\RSet$, an \emph{interval extension}
  $F\colon\ISet^n\to\ISet$ of $f$ is an interval function verifying:
\begin{align*}
    \Outer{f(\Vector{B})} & \subseteq F(\Vector{B}) & 
    \forall\Vector{B}\in\mathcal{D}^{\ISet}_f 
\end{align*}
where $\Outer{\rho}=\bigcap\{\Vector{B}\in\ISet^n\mid\rho\subseteq\Vector{B}\}$ for any real
relation $\rho\subseteq\RSet^n$.
\end{definition}

The extensions of some basic operators are defined as follows:
\begin{equation*}
  \begin{array}{ll}
    \itv{a}{b}+\itv{c}{d}=&\itv{a+c}{b+d}\\
    \itv{a}{b}-\itv{c}{d}=&\itv{a-d}{b-c}\\
    \itv{a}{b}\times\itv{c}{d}=& \itv{\min(ac,ad,bc,bd)}{\max(ac,ad,bc,bd)} \\
    \exp(\itv{a}{b}) =& \itv{\exp(a)}{\exp(b)}
  \end{array}
\end{equation*}

Note that in practice, the bounds computed have to be \emph{outward
  rounded} in order to preserve the containment property.

A particular interval extension, the \emph{natural interval extension}
is obtained by replacing syntactically in a real function all the real
constants by intervals containing them, all the real variables by
interval variables, and all operators by their interval extension.

Interval arithmetic is commutative, but it is neither associative (due
to floating-point numbers wanting for this property themselves), nor
distributive. It enjoys instead a \emph{sub-distributive property},
namely:
\begin{equation*}
  \forall I,J,K\in\ISet\colon\quad I(J+K)\subseteq IJ+IK
\end{equation*}

The sub-distributivity property has an important impact in that
equivalent forms for a function over reals may not have equivalent
natural extensions.

In the rest of this paper, we will often consider the set of real
numbers represented by a set of Cartesian product of intervals. Given
\powerset{\GSet{S}} the \emph{power set} of any set \GSet{S}, we then
introduce the operator $\BoxesToSet{\cdot}\colon\powerset{\ISet^n}\to\RSet^n$
defined as follows:
\begin{equation*}
  \forall \Vector{B_1},\dots,\forall\Vector{B_n}\in\ISet^n\colon 
  \BoxesToSet{\{\Vector{B_1},\dots,\Vector{B_n}\}}=\{\Vector{r}\in\RSet^n\mid
  \exists i\in\{1,\dots,n\}\text{ s.t. }\Vector{r}\in\Vector{B_i}\}
\end{equation*}

Lastly, given a box $\Vector{B}=I_1\times\cdots\times I_n$, an integer
$k\in\{1,\dots,n\}$, and an interval $J$, let
$\replaceDom{\Vector{B}}{I_k}{J}=I_1\times\cdots\times I_{k-1}\times
J\times I_{k+1}\times\cdots\times I_n$ be the box \Vector{B} where
$I_k$ has been replaced by $J$.

\section{Related Work}
\label{sec:related-work}

\begin{figure}
  \begin{center}
    \includegraphics[width=.5\textwidth]{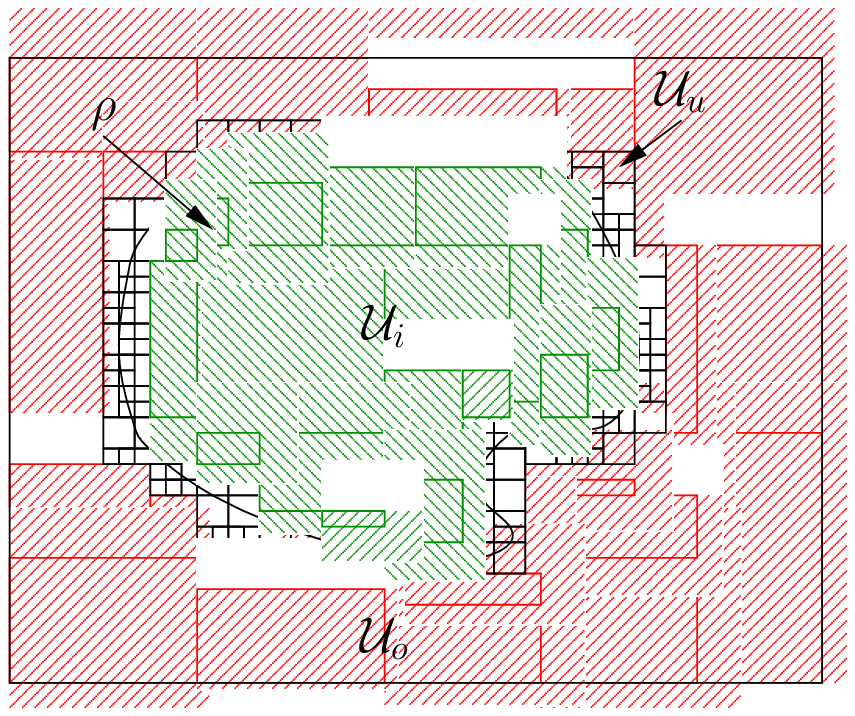}
    \caption{Subpaving of the relation $\rho$}
    \label{fig:subpaving}
  \end{center}
\end{figure}

Computing a \emph{subpaving} (Figure~\ref{fig:subpaving}) of a real
relation $\rho\subseteq\RSet^n$ consists in partitionning $\RSet^n$ into
three sets $(\GSet{U}_i,\GSet{U}_o,\GSet{U}_u)$ of non-overlapping
boxes with the properties:
\begin{equation}\label{eq:subpaving-properties}
  \left\{\begin{array}{l}
      \BoxesToSet{\GSet{U}_i}\subseteq\rho\\
      \BoxesToSet{\GSet{U}_o}\cap\rho=\emptyset\\
      \BoxesToSet{\GSet{U}_i\cup\GSet{U}_u}\supseteq\rho
      \end{array}\right.
\end{equation}

Given an $n$-ary constraint $c$, let $\CRel{c}\subseteq\RSet^n$ be the
underlying relation, that is:
\begin{equation*}
  \CRel{c}=\{(r_1,\dots,r_n)\in\RSet^n\mid c(r_1,\dots,r_n)\}
\end{equation*}

Given an $n$-ary constraint $c$ and a box $\Vector{B}\in\ISet^n$, and
assuming the existence of some procedure \algo{GlobSat} with the
following properties:
\begin{equation*}
    \left\{\begin{array}{ll}
      \algo{GlobSat}(c,\Vector{B})=\text{true} &
          \implies\Vector{B}\subseteq\CRel{c}\\
      \algo{GlobSat}(c,\Vector{B})=\text{false} &
          \implies\Vector{B}\cap\CRel{c}=\emptyset\\
      \algo{GlobSat}(c,\Vector{B})=\text{unknown} & \implies ???
      \end{array}\right.
\end{equation*}
it is easy to devise a systematic procedure to compute the subpaving
of \CRel{c} by alternating \emph{evaluation steps} with
\algo{GlobSat}, and \emph{splitting steps} whenever \algo{GlobSat}
returns ``unknown.'' The precise algorithm is presented in
Table~\ref{algo:subpaving} for a conjunction of constraints.

\begin{acmtable}{\textwidth}
\medskip
\begin{tabbing}
123\=123\=123\=123\=123\=\kill
\NUM{1} \>\algo{Subpaving}$\bigl($\kwd{in}: $\{c_1,\dots,c_m\}$, 
                 $\Vector{B}\in\ISet^n$;
                 \kwd{out}: $(\GSet{U}_i,
                       \GSet{U}_o,\GSet{U}_u)\in\powerset{\ISet^n}^3\bigr)$\\
\NUM{2}\> \BEGIN\\
\NUM{3}\> \> sat $\gets \algo{GlobSat}(c_1,\Vector{B})\wedge\cdots
                         \wedge\algo{GlobSat}(c_m,\Vector{B})$\\
\NUM{4}\> \> \SWITCH{sat} \\
\NUM{5}\> \>\> \kwd{true}: \\
\NUM{6}\> \>\>\> \RETURN{$(\{\Vector{B}\},\emptyset,\emptyset)$}\\
\NUM{7}\> \>\> \kwd{false}: \\
\NUM{8}\> \>\>\> \RETURN{$(\emptyset,\{\Vector{B}\},\emptyset)$}\\
\NUM{9}\> \>\> \kwd{unknown}: \\
\NUM{10}\> \>\>\> \IF{\algo{StoppingCriterion}$(\Vector{B})}$\\
\NUM{11}\> \>\>\>\> \RETURN{$(\emptyset,\emptyset,\{\Vector{B}\})$}\\
\NUM{12}\> \>\>\> \ELSE\\
\NUM{13}\> \>\>\>\> $(\Vector{B_1},\dots,\Vector{B_k}) \gets \algo{Split}_k(\Vector{B})$\\
\NUM{14}\> \>\>\>\> \RETURN{$\displaystyle\biguplus_{j=1}^k 
                  \algo{Subpaving}(\{c_1,\dots,c_m\},\Vector{B_j})$}\\
\NUM{15}\> \>\>\> \ENDIF\\
\NUM{16}\> \> \ENDSWITCH\\
\NUM{17}\> \END
\end{tabbing}
\caption{Subpaving algorithm for a conjunction of atomic constraints $c_1\wedge\cdots\wedge c_m$}  
\label{algo:subpaving}
\end{acmtable}

The \algo{StoppingCriterion} function appearing on Line~$10$ of
Alg.~\algo{Subpaving} returns ``true'' or ``false'' depending whether
the box given as an argument should still be considered for splitting.
A typical instance of this method is testing for canonicity of the
box. It is also possible to speed-up the computation by using another
instance of \algo{StoppingCriterion} that would avoid splitting boxes
whose width is smaller than some $\varepsilon$.

The $\algo{Split}_k$ function on Line~$13$ splits a box \Vector{B} into $k$
non-overlapping subboxes whose union is equal to \Vector{B}.

The $\biguplus$ operator on Line~$14$ applies on vectors of sets and
performs their unions componentwise:
\begin{equation*}
  \biguplus_i\biggl\{(\GSet{S}_1^i,\dots,\GSet{S}_n^i)\biggr\}=
  (\bigcup_i\GSet{S}_1^i,\dots,\bigcup_i\GSet{S}_n^i)
\end{equation*}

The \algo{subpaving} algorithm has been used by several authors to
compute sound boxes for systems of inequalities. They differ by the
way they implement the \algo{GlobSat} method.

In \algo{SIVIA}, Jaulin and Walter~\cite{Jaulin-Walter:Automatica93}
use interval arithmetic containment properties: given a constraint
$c\colon f(x_1,\dots,x_n)\leq0$, and a box \Vector{B}, they evaluate
the natural interval extension of $f$ over \Vector{B}. If the right
bound of the result is negative, \algo{GlobSat} returns
``true''; if the left bound is strictly positive \algo{GlobSat}
returns ``false''; otherwise it returns ``unknown.'' \algo{SIVIA} is
able to process any kind of inequality constraints, be they linear or
not, polynomial or not. The drawback of this approach is that for
unstable $f$ functions, the natural interval evaluation leads
to large intervals that do not permit deciding whether the box
\Vector{B} is included in $\CRel{c}$ or not. It is then necessary to 
split the box a lot.

Jaulin and Walter~\cite{Jaulin-Walter:Automatica96} have also devised
an algorithm to solve the guaranteed tuning problem (restricted to the
finding of only one value). It is based on SIVIA, and then, it suffers
from the same drawbacks as SIVIA itself.

The algorithm described by Kutsia and Schicho~\cite{Kutsia-Schicho:99}
implements another instance of \algo{Subpaving} where \algo{GlobSat}
is obtained by testing some criterion using floating-point numbers of
arbitrary precision. An important limitation is that their algorithm
can only handle polynomial strict inequalities.

Garloff and Graf~\cite{Garloff-Graf:99} also restrict themselves to
polynomial strict inequalities. They expand the polynomial inequalities into Bernstein
polynomials: let $\Vector{I}=(i_1,\dots,i_n)$ be a multi-index (vector of
non-negative integers), and $\Vector{x^{\Vector{I}}}=x_1^{i_1}\dots x_n^{i_n}$ be a
monomial. Given a polynomial $p\in\RSet[x_1,\dots,x_n]$, let $\GSet{S}$ be
a set of multi-indices such that: $p(\Vector{x})=\sum_{\Vector{I}\in\GSet{S}}a_{\Vector{I}}
\Vector{x}^{\Vector{I}}$. For any $n$-ary Cartesian product of domains \Vector{D},
one can express $p$ in terms of Bernstein coefficients:
\begin{equation*}
  p(\Vector{x})=\sum_{{\Vector{I}}\in \GSet{S}} b_{\Vector{I}}(\Vector{D})B_{N,{\Vector{I}}}(\Vector{x})
\end{equation*}
where $B_{N,{\Vector{I}}}(\Vector{x})$ is the ${\Vector{I}}$th Bernstein polynomial of degree $N$. 

It is then well known that the following property does hold~\cite{Farouki-Rajan:CAGD87}:
\begin{equation*}
  \forall\Vector{a}\in\Vector{D}\colon\min_{{\Vector{I}}\in\GSet{S}}b_{\Vector{I}}(\Vector{D})\leq
  p(\Vector{a})\leq\max_{{\Vector{I}}\in\GSet{S}}b_{\Vector{I}}(\Vector{D}) \quad
  \text{\emph{(convex hull property)}}
\end{equation*}
As a consequence, it is possible to accurately bound the range of
$p(\Vector{x})$ on every box, and then to know whether an inequality
such as $p(\Vector{x})<0$ does hold or not. By splitting the
initial box and evaluating the corresponding Bernstein coefficients,
one can isolate all the boxes verifying a set of inequality
constraints.

Sam-Haroud and Faltings~\cite{Haroud-Faltings:94} do not rely on
Alg.~\algo{Subpaving}. Instead, they represent explicitly the
subpaving of the relation \CRel{c} associated to each $n$-ary
inequality constraint by a $2^n$-tree of boxes. The category of each
box (in $\GSet{U}_i$, $\GSet{U}_o$, or $\GSet{U}_u$) is determined by
finding the intersection of the curve with the edges of the box. Sets
of constraints are handled by performing intersections of the
respective $2^n$-trees.  In order to avoid a combinatorial explosion,
$n$ must be small. As a consequence, they have chosen to decompose the
user's constraints into ternary constraints, so that they only
manipulate octrees. The original algorithm is able to compute inner
approximations (that is, subsets) of the solution space; on the other
hand, it is not designed to handle the guaranteed tuning problem.

The method presented by Collavizza et
al.~\cite{Collavizza-et-al:IJCAI99} is strongly related to the one we
present in the following, since they rely on usual interval constraint
solving techniques to compute sound boxes for some constraint system.
Starting from a seed that is known to belong to the solution space,
they enlarge the domain of the variables around it in such a way that
the new box computed is still included in the solution space. They
do so by using local consistency techniques to find the points at
which the truth value of the constraints change.  Their algorithm is
particularly well suited for the applications they target, \emph{viz.}\ the
enlargement of tolerances. It is however not designed to solve the
guaranteed tuning problem. In addition, it is necessary to obtain a
seed for each connected subset of the solution space, and to apply the
algorithm on each seed if one is interested in computing several
solutions (e.g.\ to ensure representativeness of the samples).

\begin{figure} 
\begin{center}  
  \includegraphics[width=.5\hsize]{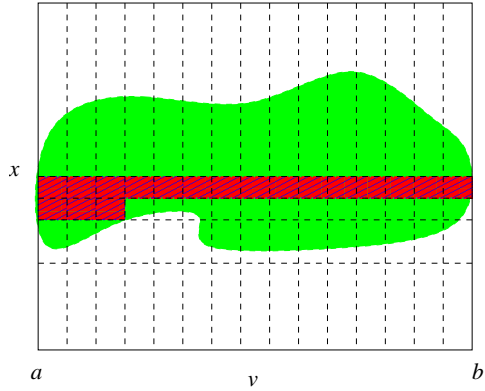}
  \caption{JLA algorithm: solving $\forall v\in\itv{a}{b}\colon c(x,v)$}
  \label{fig:jla} 
\end{center}
\end{figure}

In order to tighten a box \Vector{B} of variables' domains for a
problem of the form $\forall v\in I_k\colon c_1\wedge\cdots\wedge c_m$,
Jardillier and Langu\'enou~\cite{Jardillier-Languenou:EG98} compute an
inner approximation by decomposing the initial domain $I_k$ of $v$
into canonical intervals $I_k^1$,\dots,$I_k^p$, and testing whether
$c_1\wedge\cdots\wedge c_m$ does hold for the boxes
$I_1\times\cdots\times I_k^1\times\cdots\times I_n$,\dots,
$I_1\times\cdots\times I_k^p\times\cdots\times I_n$.  These
evaluations give results in a three-valued logic (\emph{true},
\emph{false}, \emph{unknown}). Boxes labeled \emph{true} contain only
solutions, boxes labeled \emph{false} contain no solution at all, and
boxes labeled \emph{unknown} are recursively split and re-tested until
they may be asserted true or false, or canonicity is reached (see Figure~\ref{fig:jla}).
Retained boxes are those verifying:
\begin{equation*}\label{eq:retained-equations}
\forall j\in\{1,\dots,p\}\colon
\left\{\begin{array}{l}
\algo{GlobSat}(c_1,I_1\times\cdots\times I_k^j\times\cdots\times I_n)=\text{true}\\
\wedge\\
\dots\\
\wedge\\
\algo{GlobSat}(c_m,I_1\times\cdots\times I_k^j\times\cdots\times I_n)=\text{true}
\end{array}\right.
\end{equation*}

\begin{acmtable}{\textwidth}
\medskip
\begin{tabbing}
123\=123\=123\=123\=123\=123\=123\=\kill
\NUM{1} \>\algo{JLA}$\bigl($\kwd{in}: $\{c_1,\dots,c_m\}$, 
                 $\Vector{B}\in\ISet^n$, a variable $v$, $l\in\FSet$;
                 \kwd{out}: $(\GSet{U}_i,
                       \GSet{U}_o,\GSet{U}_u)\in\powerset{\ISet^n}^3\bigr)$\\
\NUM{2}\> \BEGIN\\
\NUM{3}\> \> $\Vector{B'}\gets\replaceDom{\Vector{B}}{I_v}{\itv{l}{\nextFloat{l}}}$\\
\NUM{4}\> \> sat $\gets \algo{GlobSat}(c_1,\Vector{B})\wedge\cdots
                         \wedge\algo{GlobSat}(c_m,\Vector{B})$\\
\NUM{5}\> \> \SWITCH{sat} \\
\NUM{6}\> \>\> \kwd{true}: \\
\NUM{7}\> \>\>\> \IF{$\nextFloat{l}=\RB{v}$}\\
\NUM{8}\> \>\>\>\> \RETURN{$(\{\Vector{B}\},\emptyset,\emptyset)$}\\
\NUM{9}\> \>\>\> \ELSE\\
\NUM{10}\> \>\>\>\> \RETURN{\algo{JLA}$(\{c_1,\dots,c_m\},
                                            \Vector{B},v,\nextFloat{l})$}\\
\NUM{11}\> \>\>\> \ENDIF\\
\NUM{12}\> \>\> \kwd{false}: \\
\NUM{13}\> \>\>\> \RETURN{$(\emptyset,\{\Vector{B}\},\emptyset)$}\\
\NUM{14}\> \>\> \kwd{unknown}: \\
\NUM{15}\> \>\>\> \IF{$\algo{StoppingCriterion}(\Vector{B})$}\\
\NUM{16}\> \>\>\>\> \RETURN{$(\emptyset,\emptyset,\{\Vector{B}\})$}\\
\NUM{17}\> \>\>\> \ELSE\\
\NUM{18}\> \>\>\>\> $(\Vector{B_1},\dots,\Vector{B_k}) \gets \algo{Split}^{\setminus v}_k(\Vector{B})$\\
\NUM{19}\> \>\>\>\> \RETURN{$\displaystyle\biguplus_{j=1}^k 
                  \algo{JLA}(\{c_1,\dots,c_m\},\Vector{B_j},v,l)$}\\
\NUM{20}\> \>\>\> \ENDIF\\
\NUM{21}\> \> \ENDSWITCH\\
\NUM{22}\> \END
\end{tabbing}
\caption{Evaluation-based propagation algorithm for  
  $\forall v\in\itv{\LB{v}}{\RB{v}}\colon c_1\wedge\cdots\wedge c_m$} 
\label{algo:JLA}
\end{acmtable}

The precise algorithm is described in Table~\ref{algo:JLA}. In the
initial call, the $l$ parameter is equal to the left bound of $v$'s
domain. The $\algo{Split}^{\setminus v}_k$
procedure in Line~$18$ is identical to the $\algo{Split}_k$ procedure
presented previously, except that it never splits the domain of the
universally quantified variable $v$.

Once again, Alg.~\algo{JLA} is but a sophisticated instance of
Alg.~\algo{Subpaving}, which means that its efficiency strongly
depends on the quality of the \algo{GlobSat} procedure. Like in SIVIA,
Jardillier and Langu\'enou use the natural interval extension of the
constraints. As a consequence, their algorithm is computationally
expensive in many cases, as soon as some moderate precision in the
computation of the inner approximation is required. We refer the reader to 
Section~\ref{sec:benchmarking} for precise figures.

Due to lack of space, we will not present the Cylindrical Algebraic
Decomposition method for quantifier elimination here, and we refer the
reader to the good introduction by Jirstrand~\cite{Jirstrand:RR95}.

\section{Interval Constraint Solving}
\label{sec:interval-constraint-solving}

Since finite representation of numbers by computers hinders the
reliable solving of real equations and inequations, \emph{interval
  constraint solving} relies on \emph{interval
  arithmetic}~\cite{Moore:66,Alefeld-Herzberger:83} to compute
verified approximate solutions of real constraint systems.  

The reader will find in various
works~\cite{Older-Vellino:CCECE90,Benhamou-Older:JLP97,Benhamou:ALP96,Van-Hentenryck-et-al:97,Benhamou-et-al:ILPS94,Van-Hentenryck-et-al:JNA97,Benhamou-et-al:ICLP99}
a thorough presentation of the interval constraint solving framework
and of the associated state-of-the-art algorithms. In this section, we
will describe only the elements that are essential to our purpose. In
particular, we will restrict ourselves to the case of nonlinear
inequality constraints only, that is constraints of the form
$f(x_1,\dots,x_n)\leq0$ for some real function $f$.

Proofs not given here may be found in the papers cited above.

\subsection{Local Consistencies}
\label{sec:local-consistencies}

Discarding all inconsistent values from a box of variables' domains is
intractable when the constraints are real ones (consider
for instance the constraint $c\colon\sin(x)=1,\quad x\in\itv{0}{2}$).
Consequently, weak consistencies have been devised, among which one may
cite \emph{hull consistency}~\cite{Benhamou:Chatillon94} and \emph{box
  consistency}~\cite{Benhamou-et-al:ILPS94}. Both consistencies permit narrowing
variables' domains to (hopefully) smaller domains, preserving all the
solutions present in the input. Since they are used as a basis for the
algorithms to be introduced in Section~\ref{sec:sound-constraint-solving}, they are
both presented below.

Discarding values of the variable domains for which $c$ does not hold
according to a given consistency notion is modelled by means of outer contracting
operators, whose main properties are \emph{contractance},
\emph{completeness}, and \emph{monotonicity}.

\subsubsection{Outer Contracting Operators}
\label{sec:contracting-operators}

Depending on the considered consistency, one may define different
contracting operators for a constraint. In this section, hull
consistency and box consistency are first formally presented. 
Operators based on both consistencies are then given.

Hull consistency is a strong ``weak consistency'' since a constraint
$c$ is said hull consistent w.r.t.\ a box whenever that box cannot be
tightened without losing some solutions for $c$:

\begin{definition}[Hull consistency]
  A real constraint $c(x_1,\dots,x_n)$ is said \emph{
    hull-consistent w.r.t.\ a box} $\Vector{B}=I_1\times\cdots\times I_n$ 
  if and only if:
  \begin{multline*}
    \forall k\in\{1,\dots,n\}\colon \\
    I_k\mathrel=\Outer{I_k\cap
    \{r_k\in\RSet\mid\forall j\in\{1,\dots,n\}\setminus\{k\}\colon\\
    \exists r_j\in I_j \text{ s.t. }  
    (r_1,\dots,r_n)\in\CRel{c}\}}
  \end{multline*}
\end{definition}

An operator enforcing hull consistency for a constraint $c$
computes the smallest box containing the intersection of the
input box and the relation \CRel{c}:

\begin{definition}[Outer-hull contracting operator]
  \label{def:outer-hull-contracting-operator-definition}
  Let $c$ be an $n$-ary constraint, \CRel{c} its underlying relation,
  and $\Vector{B}$ a box.  An \emph{outer-hull contracting operator}
  for $c$ is a function $\algo{HC}_{c}\colon\ISet^n\to\ISet^n$ defined
  by:
  \begin{equation*}
    \algo{HC}_c(\Vector{B})=\Outer{\Vector{B}\cap\CRel{c}}
  \end{equation*}
\end{definition}

\begin{proposition}[Completeness of \algo{HC}]
  \label{prop:completeness-OC}
  Given a constraint $c$ and a box \Vector{B}, the following relation
  does hold: $(\Vector{B}\cap\CRel{c})\subseteq\algo{HC}_c(\Vector{B})$.
\end{proposition}

Operationally, hull consistency is enforced over constraints by
decomposing them into conjunctions of \emph{primitive constraints}.
The drawback of such a method lies in the loss of domain tightening
due to the introduction of new variables during the decomposition
process.  Box consistency has been introduced by Benhamou
et al.~\cite{Benhamou-et-al:ILPS94} to overcome this problem: the operators
enforcing box consistency consider constraints globally, without
decomposing them. 

\begin{definition}[Box consistency]
  Let $c$ be an $n$-ary real constraint, $C$ an interval extension for $c$, and
  $\Vector{B}=I_1\times\cdots\times I_n$ a box. The constraint $c$ is
  said \emph{box-consistent w.r.t.\ \Vector{B}} if and only if:
  \begin{multline*}
    \forall k\in\{1,\dots,n\}\colon\\
     I_k=\Outer{I_k\cap\{r\in\RSet\mid
       C(I_1,\dots,I_{k-1},\Outer{\{r\}},I_{k+1},\dots,I_n)\}}
  \end{multline*}
\end{definition} 

Intuitively, a constraint $c$ is box-consistent w.r.t.\ a box \Vector{B}
when each projection $I_j, j\in\{1,\dots,n\}$ of \Vector{B} is the
smallest interval containing all the elements that cannot be
distinguished from solutions of the univariate constraint obtained from $C$ 
by replacing all the variables but $x_j$ by their current domain due to
the inherently limited precision of the computation with
floating-point numbers.

Box consistency is enforced over a constraint $c(x_1,\dots,x_n)$ as
follows: $n$ contracting operators (typically using an interval Newton
method~\cite{Van-Hentenryck-et-al:97}) are associated to the $n$
univariate interval constraints obtained as described above.  The $k$th
operator reduces the domain of $x_k$ by computing the leftmost
and rightmost canonical intervals $J$ such that
$C(I_1,\dots,I_{k-1},J,I_{k+1},\dots,I_n)$ does hold (\emph{leftmost
  and rightmost quasi-zeros}).

Using box consistency to narrow down the variable domains of a
constraint leads to the notion of \emph{outer-box contracting operator}:

\begin{definition}[Outer-box contracting operator]
\label{def:outer-box-operator}
  Given an $n$-ary constraint $c$ and a box $\Vector{B}$, an
  \emph{outer-box contracting operator}
  $\algo{BC3r}_{c}\colon\ISet^n\to\ISet^n$ for $c$ is defined
  by:
  \begin{multline*}
    \algo{BC3r}_c(\Vector{B})=\max\{\Vector{B'}\mid 
    \Vector{B'}\subseteq\Vector{B}\\
    \text{ and } 
    c \text{ is box-consistent w.r.t. } k\in\{1,\dots,n\} \text{ and } 
    \Vector{B'}\}
  \end{multline*}
\end{definition}

\begin{proposition}[Completeness of \algo{BC3r}]
  Given a constraint $c$, the following relation does hold for any box
  \Vector{B}:
\begin{equation*}
  (\Vector{B}\cap\CRel{c})\subseteq\algo{BC3r}_c(\Vector{B})
\end{equation*}
\end{proposition}

Consistencies and associated contracting operators considered
so far are such that completeness is guaranteed (no solution lost
during the narrowing process). Devising operators that ensure soundness
of the results is the topic of the next section.

\section{Sound Interval Constraint Solving}
\label{sec:sound-constraint-solving}

We have seen in the previous section some operators that compute an
outer approximation of the relation $\rho$ associated to some
nonlinear constraint system. In this section, we will present
operators that compute an \emph{inner approximation} of $\rho$, that
is, a subset of the solution space of the corresponding constraint
system. As a consequence, the boxes computed by these operators will
have the property that any point inside of them is a true solution to
the problem, thus ensuring soundness of the results given to the user.

Several definitions for the inner approximation of a real relation
exist in the literature, depending on the intended application. Given
an $n$-ary relation $\rho$, one may single out at least the two following
definitions for an inner approximation $\Inner{\rho}$ of
$\rho$:
\begin{enumerate}
\item \textbf{Def.\ \textbf{A.}}  $\Inner{\rho}=\Vector{B_1}$,
  where
  $\Vector{B_1}\in\{\Vector{B}\in\ISet^n\mid\Vector{B}\subseteq\rho\}$
  (an inner-approximation is any box included in the
  relation)~\cite{Markov:JUCS95,Armengol-et-al:CCIA98};
\item \textbf{Def.\ \textbf{B.}}
  $\Inner{\rho}=\Vector{B_1}$, where
  $\Vector{B_1}\in\{\Vector{B}=I_1\times\cdots\times I_n\mid
  \Vector{B}\subseteq\rho\wedge \forall j\in\{1,\dots,n\},\forall
  I'_j\supseteq I_j\colon
  \replaceDom{\Vector{B}}{I_j}{I'_j}\subseteq\rho\implies
  I'_j=I_j\}$ (an inner-approximation is a box included in the
  relation that cannot be extended in any direction without containing
  non-solution points)~\cite{Shary:MISC99}.
\end{enumerate}

Definitions~\textbf{A} and \textbf{B} imply that disconnected
relations are only very partially represented by one box only, a
drawback that is avoided with the following stronger definition, which will
be used in this paper:

\begin{definition}[Inner approximation operator]
  \label{def:inner-approximation}
  Given an $n$-ary real relation $\rho\subseteq\RSet^n$, the inner approximation
  operator $\algo{Inner}\colon\allowbreak\RSet^n\to\RSet^n$ is defined by:
  \begin{equation*}
    \Inner{\rho}=\{\Vector{r}\in\RSet^n \mid
    \Outer{\{\Vector{r}\}}\subseteq\rho\}
  \end{equation*}
\end{definition}

The inner-approximation contains all the elements whose enclosing box
is included in the relation. The \algo{Inner} operator enjoys the
following properties:

\begin{proposition}[Properties of the \algo{Inner} operator]
  \label{prop:properties-Inner}
  The \algo{Inner} operator is contracting, monotonic, idempotent,
  sub-distributive w.r.t.\ the union of subsets of $\RSet^n$, and 
  distributive w.r.t.\ the intersection of subsets of $\RSet^n$.
  
  \begin{proof}
    Given \GSet{A}, \GSet{B}, and \GSet{C} three subsets of $\RSet^n$, with 
    $\GSet{A}\subseteq\GSet{B}$, let
    \Vector{r} be an element of
    $\RSet^n$ and $\Vector{D}=\Outer{\{\Vector{r}\}}$.
    \begin{description}
    \item [Contractance] Given any real $\Vector{r}\in\Inner{\rho}$, we have by
      definition of \algo{Inner} that $\Outer{\{\Vector{r}\}}\subseteq\rho$. By
      definition of \algo{Outer}, we have that $\Vector{r}\in\Outer{\{\Vector{r}\}}$,
      which leads to $\Vector{r}\in\rho$. Consequently,
      $\Inner{\rho}\subseteq\rho$;
    \item [Monotonicity] For any $\Vector{r}$,
      $\Vector{r}\in\Inner{\GSet{A}}$ implies
      $\Vector{D}\subseteq\GSet{A}$, by definition of \algo{Inner};
      since $\GSet{A}\subseteq\GSet{B}$, we have
      $\Vector{D}\subseteq\GSet{B}$, and finally, by definition of
      \algo{Inner}, $\Vector{r}\in\Inner{\GSet{B}}$. Consequently,
      $\GSet{A}\subseteq\GSet{B}\implies\Inner{\GSet{A}}\subseteq\Inner{\GSet{B}}$;
    \item [Idempotence] We first prove
      $\Inner{\Inner{\GSet{A}}}\subseteq\Inner{\GSet{A}}$: \par 
      Given
      $\Vector{r}\in\Inner{\Inner{\GSet{A}}}$, we have
      $\Vector{D}\subseteq\Inner{\GSet{A}}$, by definition of
      \algo{Inner}.  By contractance of \algo{Inner},
      $\Inner{\GSet{A}}\subseteq\GSet{A}$. We then have
      $\Vector{D}\subseteq\GSet{A}$, and then
      $\Vector{r}\in\Inner{\GSet{A}}$, by definition of \algo{Inner}
      again.\par
      Now, we prove $\Inner{\Inner{\GSet{A}}}\supseteq\Inner{\GSet{A}}$: \par
      Given $\Vector{r}\in\Inner{\GSet{A}}$, we have 
      $\Vector{D}\subseteq\GSet{A}$ by definition of \algo{Inner}. Therefore, 
      any element in \Vector{D} is in \Inner{\GSet{A}}. As a consequence,
      $\Vector{D}\subseteq\Inner{\GSet{A}}$. Definition of \algo{Inner} then leads to
      $\Vector{r}\in\Inner{\Inner{\GSet{A}}}$;
    \item [Sub-distributivity w.r.t.\ union] 
      We prove that: 
      \begin{equation*}
        \Inner{\GSet{B}}\cup\Inner{\GSet{C}}\subseteq
        \Inner{\GSet{B}\cup\GSet{C}}
      \end{equation*}
      We have:
      \begin{align*}
        \Inner{\GSet{B}}\cup\Inner{\GSet{C}} &
        = \bigl\{r\in\RSet^n\mid\Outer{\{r\}}\subseteq\GSet{B}\bigr\}\cup
\bigl\{r\in\RSet^n\mid\Outer{\{r\}}\subseteq\GSet{C}\bigr\}\\
        & =\bigl\{r\in\RSet^n\mid\Outer{\{r\}}\subseteq\GSet{B}\vee\Outer{\{r\}}\subseteq\GSet{C}\bigr\}\\
        &\subseteq\bigl\{r\in\RSet^n\mid\Outer{\{r\}}\subseteq\GSet{B}\cup\GSet{C}\bigr\}\\
        &\subseteq\Inner{\GSet{B}\cup\GSet{C}}
      \end{align*}
    \item [Distributivity w.r.t.\ intersection]
      We prove that: 
      \begin{equation*}
        \Inner{\GSet{B}}\cap\Inner{\GSet{C}}=
        \Inner{\GSet{B}\cap\GSet{C}}
      \end{equation*}
      We have:
      \begin{align*}
        \Inner{\GSet{B}\cap\GSet{C}} &=\bigl\{r\in\RSet^n\mid\Outer{\{r\}}\subseteq(\GSet{B}\cap\GSet{C})\bigr\}\\
          & = \bigl\{r\in\RSet^n\mid\Outer{\{r\}}\subseteq\GSet{B}\wedge\Outer{\{r\}}\subseteq\GSet{C}\bigr\}\\
          &= \bigl\{r\in\RSet^n\mid\Outer{\{r\}}\subseteq\GSet{B}\bigr\}\cap\bigl\{r\in\RSet^n\mid\Outer{\{r\}}\subseteq\GSet{C}\bigr\}\\
          &=\Inner{\GSet{B}}\cap\Inner{\GSet{C}}
      \end{align*}
    \end{description}
  \end{proof}
\end{proposition}

\subsection{Inner Contracting Operators}
\label{sec:inner-contracting-operator}

The narrowing of variable domains occurring in a constraint is done in
the same way as in the outer-approxima\-tion case: an
\emph{inner contracting operator} associated to each constraint
discards from the initial box all the inconsistent values along with
consistent values that cannot be enclosed in a canonical
computer-representable box. The result is a set of boxes.

\begin{definition}[Inner contracting operator]
  Let $c$ be an $n$-ary constraint. An
  \emph{inner contracting operator} for $c$ is a function
  $\algo{IC}_{c}\colon\ISet^n\to\powerset{\ISet^n}$ verifying:
  \begin{equation*}
    \forall\Vector{B}\in\ISet^n\colon
    \BoxesToSet{\algo{IC}_c(\Vector{B})}\subseteq\Inner{\Vector{B}\cap\CRel{c}}
  \end{equation*}
\end{definition}

\begin{proposition}[Soundness of \algo{IC}]
  \label{prop:correctness-IC}
Given a constraint $c$ and an inner-con\-tra\-c\-ting operator
  $\algo{IC}_c$ for $c$, we have:
  \begin{equation*}
    \forall\Vector{B}\in\ISet^n\colon
    \BoxesToSet{\algo{IC}_c(\Vector{B})}\subseteq(\Vector{B}\cap\CRel{c})
  \end{equation*}
  \begin{proof}
    Immediate consequence of \algo{Inner} and \algo{Outer} definitions.
  \end{proof}
\end{proposition}

\begin{remark}
  Given a constraint $c$, an inner contracting operator $\algo{IC}_c$
  for $c$, an outer contracting operator $\algo{OC}_c$ for $c$,
  and a box \Vector{B}, the following relations, deriving from
  Prop.~\ref{prop:completeness-OC} and
  Prop.~\ref{prop:correctness-IC}, hold:
  \begin{equation*}
    \BoxesToSet{\algo{IC}_c(\Vector{B})}\subseteq(\Vector{B}\cap\CRel{c})
    \subseteq\algo{OC}_c(\Vector{B})
  \end{equation*}
\end{remark}

Inner-contracting operators with stronger properties (computation of
the greatest representable set included in a relation) may be defined,
provided some strong assumption (namely the ability to determine for any
canonical box whether it is included or not in the relation), is
fulfilled. These operators are \emph{optimal} in the sense defined
below.

\begin{definition}[Optimal inner contracting operator]
Let $c$ be an $n$-ary constraint, and
  $\algo{IC}_c$ an inner-contracting operator for $c$. The operator
  $\algo{IC}_c$ is said \emph{optimal} if and only if the following
  relation does hold for any box \Vector{B}:
  \begin{equation*}\label{eq:definition-optimal-IC}
    \BoxesToSet{\algo{IC}_c(\Vector{B})}=\Inner{\Vector{B}\cap\CRel{c}}
  \end{equation*}
\end{definition}

Let us consider the following naive algorithm as an implementation of
an optimal inner contracting operator: given a constraint $c$, and a
box \Vector{B} of domains for the variables occurring in $c$, let us
split \Vector{B} in all its canonical subboxes. We can apply the
outer-hull contracting operator $\algo{HC}_c$ enforcing hull consistency
on each of these canonical subboxes. Consider the possible cases
for one of the canonical subboxes \Vector{D}:
\begin{itemize}
\item[$\diamond$] $\Vector{D}\cap\CRel{c}=\emptyset$. By definition of hull consistency,
  we must have $\algo{HC}_c(\Vector{D})=\emptyset$;
\item[$\diamond$] $\Vector{D}\cap\CRel{c}\neq\emptyset$. There are again several 
  cases (considered exclusively, from top to bottom):
  \begin{itemize}
  \item[$\cdot$] $\Vector{D}\subseteq\CRel{c}$: By completeness of hull
    consistency-based operators, we must have
    $\algo{HC}_c(\Vector{D})=\Vector{D}$. Now, if we consider the
    negation \negation{c} of $c$, we must have
    $\algo{HC}_{\negation{c}}(\Vector{D})=\emptyset$, by definition of
    hull consistency, since \Vector{D} does not contain any element of
    \CRel{\negation{c}},
  \item[$\cdot$] $\Vector{D}\cap\CRel{\negation{c}}\neq\emptyset$. The
    box \Vector{D} contains both some elements of \CRel{c} and
    \CRel{\negation{c}}.  Consequently, by completeness,
    $\algo{HC}_c(\Vector{D})$ and
    $\algo{HC}_{\negation{c}}(\Vector{D})$ must return non-empty boxes 
    included in \Vector{D} (a canonical box may still be tightened
    by shrinking any of its non-punctual dimensions to one of the bounds).
  \end{itemize}
\end{itemize}

We then have a procedure to determine for each canonical box
\Vector{D} whether it is part of \Inner{\CRel{c}} or not, namely:
\begin{itemize}
\item[$\diamond$] $\algo{HC}_c(\Vector{D})=\emptyset\implies\Vector{D}\not\subseteq\Inner{\CRel{c}}$;
\item[$\diamond$] $\algo{HC}_{\negation{c}}(\Vector{D})=\emptyset\implies
  \Vector{D}\subseteq\Inner{\CRel{c}}$;
\item[$\diamond$] $\bigl(\algo{HC}_c(\Vector{D})\subseteq\Vector{D}\wedge
  \algo{HC}_{\negation{c}}(\Vector{D})\subseteq\Vector{D}\bigr) \implies 
  \Vector{D}\not\subseteq\Inner{\CRel{c}}$.
\end{itemize}

By contractance and completeness of $\algo{HC}_c$, there are no other
possible outcomes (note that all cases are considered mutually
exclusive for any non-empty canonical interval).

Now, the preceding rules do not allow to implement an optimal
inner contracting operator in practice for the following reasons
(disregarding the fact that it would be grossly inefficient anyway):
\begin{itemize}
\item We have restricted our presentation of consistency operators to
  closed interval arithmetic only. As a consequence, for any
  constraint of the form $f(x_1,\dots,x_n)\leq0$, we cannot
  consider its negation $f(x_1,\dots,x_n)>0$, which would require open
  interval arithmetic as well. Instead, we will define its negation as
  the constraint $f(x_1,\dots,x_n)\geq0$. As a result, the property
  ``$\algo{HC}_{\negation{c}}(\Vector{D})\neq\emptyset\implies\Vector{D}\not\subseteq\Inner{\CRel{c}}$''
  does not hold any more. Consider for example the constraint $c\colon
  x\geq0.5$ with $x\in\itv{0.5}{\nextFloat{0.5}}$. We have
  $\algo{HC}_c(\itv{0.5}{\nextFloat{0.5}})=\itv{0.5}{\nextFloat{0.5}}$;
  considering $\negation{c}$ as the constraint $x\leq0.5$, we obtain:
  $\algo{HC}_{\negation{c}}(\itv{0.5}{\nextFloat{0.5}})=\itv{0.5}{0.5}$.
  Applying the rules above, we would conclude that the interval
  \itv{0.5}{\nextFloat{0.5}} does not belong to \Inner{\CRel{c}},
  though it does. It is however important to note that the ``relaxed''
  definition for the negation we are forced to adopt has no effect on
  the soundness of the algorithms to be described. On the other hand,
  it will preclude us from devising optimal inner contractors;
\item Even if we were using open interval arithmetic, we still would have to
  face the fact that we are usually not able to implement operators
  enforcing hull consistency on the constraints given by the user but
  on primitives obtained after the decomposition process;
\item Lastly, we have seen in Section~\ref{sec:interval-analysis} that the
  correct rounding of most floating-point arithmetic operators is
  usually not guaranteed, which precludes us from implementing contracting operators
  that enforce hull consistency, even for primitive constraints.
\end{itemize}

Nevertheless, we can still use this principle to implement non-optimal
inner contracting operators based on any kind of outer contracting
operator. Table~\ref{algo:ICO1} presents such an implementation. The
operator \algo{ICO1} is an inner contracting operator for a constraint
$c$ parameterized by an outer contracting operator $\Gamma$ for $c$.

\begin{acmtable}{\textwidth}
\medskip
\begin{tabbing}
123\=123\=123\=123\=123\=\kill
\NUM{1} \>$\algo{ICO1}^\Gamma_c\bigl(\kwd{in}\colon \Vector{B}\in\ISet^n$;
        $\kwd{out}\colon (\GSet{U}_i,\GSet{U}_o,\GSet{U}_u)\in\powerset{\ISet^n}^3\bigr)$\\
\NUM{2}\> \BEGIN\\
\NUM{3}\> \> $\Vector{B'}\gets \Gamma_c(\Vector{B})$ \\
\NUM{4}\> \> $\GSet{U}_o \gets \BComplement{\Vector{B}}{\Vector{B'}}$\\
\NUM{5}\> \> $\Vector{B''}\gets \Gamma_{\negation{c}}(\Vector{B'})$ \\
\NUM{6}\> \> $\GSet{U}_i \gets \BComplement{\Vector{B'}}{\Vector{B''}}$\\
\NUM{7}\> \> \IF{$\algo{StoppingCriterion}(\Vector{B''})$}\\
\NUM{8}\> \> \> \RETURN{$(\GSet{U}_i,\GSet{U}_o,\{\Vector{B''}\})$}\\
\NUM{9}\> \> \ELSE\\
\NUM{10}\> \>\> $(\Vector{B_1},\dots,\Vector{B_k})\gets\algo{Split}_k(\Vector{B''})$\\
\NUM{11}\> \>\> \RETURN{$(\GSet{U}_i,\GSet{U}_o,\emptyset)\uplus
                  \biggl(\smash{\displaystyle\biguplus_{j=1}^k} 
                          \algo{ICO1}^\Gamma_c(\Vector{B_j})\biggr)$}\\
\NUM{12}\> \> \ENDIF\\
\NUM{13}\> \END
\end{tabbing}
\caption{Inner contracting operator for an atomic constraint $c$ based on the outer 
  contracting operator $\Gamma$}
\label{algo:ICO1}
\end{acmtable}

The operator $\boxbslash$ occurring on lines $4$ and $6$ returns the set difference between 
two boxes as a set of boxes, that is:
\begin{multline*}
  \forall\Vector{B_1},\Vector{B_2}\in\ISet^n\colon
  \BComplement{\Vector{B_1}}{\Vector{B_2}}\subseteq\powerset{\ISet^n}\\
  \text{with}\quad
  \BoxesToSet{\BComplement{\Vector{B_1}}{\Vector{B_2}}}=
  \{r\in\Vector{B_1}\mid\Outer{\{r\}}\cap\Vector{B_2}=\emptyset\}
\end{multline*}
The result is not uniquely defined since there are many ways to represent a set by
unions of boxes. This is however irrelevant to our purpose.

\begin{proposition}[Soundness of \algo{ICO1}]
  Given an $n$-ary constraint $c$, a box $\Vector{B}\in\ISet^n$, an
  outer contracting operator $\Gamma$ for $c$, and
  $(\GSet{U}_i,\GSet{U}_o,\GSet{U}_u)=\algo{ICO1}_c^\Gamma(\Vector{B})$,
  we have:
\begin{equation*}
  \left\{\begin{array}{ll}
      \forall\Vector{D}\in\GSet{U}_o\colon& \Vector{D}\cap\CRel{c}=\emptyset\\
      \forall\Vector{D}\in\GSet{U}_i\colon& \Vector{D}\subseteq\CRel{c}
      \end{array}\right.
\end{equation*}
\begin{proof}
  Consider Line $3$ of \algo{ICO1}. By completeness of $\Gamma$,
  $\Vector{B}\cap\CRel{c}=\Vector{B'}\cap\CRel{c}$. Consequently,
  $\BoxesToSet{\BComplement{\Vector{B}}{\Vector{B'}}}\cap\CRel{c}=\emptyset$.
  Since the set $\GSet{U}_o$ returned eventually is the union of all
  the sets computed on Line $4$, we have that
  $\GSet{U}_o\cap\CRel{c}=\emptyset$. We can use the dual reasoning to
  prove that
  $\forall\Vector{D}\in\GSet{U}_i\colon\Vector{D}\subseteq\CRel{c}$.
\end{proof}
\end{proposition}

\subsection{Solving the Unidimensional Guaranteed Tuning Problem}
\label{sec:unidimensional-GTP}

We now consider the Unidimensional Guaranteed Tuning Problem (UGTP): given a set of
nonlinear inequalities $c_1,\dots,c_m$ on $n$ variables $\{x_1,\dots,x_{n-1},v\}$, a box 
$\Vector{B}\in\ISet^n$, and a domain $J$, compute the set \GSet{I} defined by:
\begin{equation}
\label{eq:unidimensional-GTP}
  \GSet{I}=\{\Vector{r}\in\Vector{B}\mid
      \forall v\in J\colon c_1\wedge\cdots\wedge c_m\}
\end{equation}

If we restrict ourselves to the case $m=1$, it is easy to devise an
algorithm implementing an inner contracting operator computing a
subset of \GSet{I} along the same lines as \algo{ICO1}: given the
constraint $\forall v\in J\colon c$, use \algo{ICO1} to compute an
inner approximation of \CRel{c} while retaining only the boxes for
which the domain of $v$ is equal to $J$. The modified algorithm is
presented in Table~\ref{algo:ICO2}. The algorithm \algo{ICO2} is
parameterized by an outer contracting operator $\Gamma$ for $c$, by
the variable $v$ that is universally quantified, and by the domain $J$
on which $v$ must range.

\begin{acmtable}{\textwidth}
\medskip
\begin{tabbing}
123\=123\=123\=123\=123\=\kill
\NUM{1} \>$\algo{ICO2}^{\Gamma,(v,J)}_c\bigl(\kwd{in}\colon \Vector{B}\in\ISet^n$; 
        $\kwd{out}\colon (\GSet{U}_i,\GSet{U}_o,\GSet{U}_u)\in\powerset{\ISet^n}^3\bigr)$\\
\NUM{2}\> \BEGIN\\
\NUM{3}\> \> $\Vector{B'}\gets \Gamma_c(\Vector{B})$ \\
\NUM{4}\> \> \IF{$\Domain{v}{\Vector{B'}} = \Domain{v}{\Vector{B}}$}\\
\NUM{5}\> \>\> $\GSet{U}_o \gets \BComplement{\Vector{B}}{\Vector{B'}}$\\
\NUM{6}\> \>\> $\Vector{B''}\gets \Gamma_{\negation{c}}(\Vector{B'})$ \\
\NUM{7}\> \>\> $\GSet{U}_i \gets \BComplement{\replaceDom{\Vector{B'}}{\Domain{v}{\Vector{B'}}}{J}}{\replaceDom{\Vector{B''}}{\Domain{v}{\Vector{B''}}}{J}}$\\
\NUM{8}\> \>\> \IF{$\algo{StoppingCriterion}(\Vector{B''})$}\\
\NUM{9}\> \>\>\> \RETURN{$(\GSet{U}_i,\GSet{U}_o,\{\Vector{B''}\})$}\\
\NUM{10}\> \>\> \ELSE\\
\NUM{11}\> \>\>\> $(\Vector{B_1},\dots,\Vector{B_k})\gets
                          \algo{Split}^{\setminus v}_k(\Vector{B''})$\\
\NUM{12}\> \>\>\> $(\GSet{U}_i,\GSet{U}_o,\GSet{U}_u)\gets 
                  (\GSet{U}_i,\GSet{U}_o,\emptyset)\uplus
                  \biggl(\smash{\displaystyle\biguplus_{j=1}^k} 
                          \algo{ICO2}^{\Gamma,(v,J)}_c(\Vector{B_j})\biggr)$\\
\NUM{13}\> \>\>\> \RETURN{$(\GSet{U}_i,\GSet{U}_o,\GSet{U}_u)$}\\
\NUM{14}\> \>\> \ENDIF\\
\NUM{15}\> \> \ELSE\\
\NUM{16}\> \>\> \RETURN{$(\emptyset,\{\Vector{B}\},\emptyset)$}\\
\NUM{17}\> \> \ENDIF\\
\NUM{18}\> \END
\end{tabbing}
\caption{Inner contracting operator for a constraint 
  $\forall v\in J\colon c$}
\label{algo:ICO2}
\end{acmtable}

The operator \Domain{v}{\Vector{B}} occurring on Line~$4$ returns the
interval in box \Vector{B} that represents the domain of $v$.  Note
that after applying $\Gamma_c$ on Line~$3$, we check that the domain
of $v$ has not been tightened. Otherwise, there would exist some value
in the domain of $v$ such that $c$ does not hold. The box \Vector{B}
would then have to be discarded. The domain of $v$ may however be
tightened on Line~$6$ from some domain $J_a$ to some domain $J_b$ by
the outer contracting operator for \negation{c}.  This corresponds to
having proved that $\forall x_1\in I'_1\setminus I''_1\dots\forall
x_{n-1}\in I'_{n-1}\setminus I''_{n-1}\forall v\in J_a\setminus
J_b\colon c$. It is then sufficient to prove $\forall v\in J_b\colon
c$ in the next steps of the algorithm.

In the following, given an $n$-ary constraint $c$, an interval $J$, and a
variable $v$, let us note \CRel{c^\forall} the $n$-ary relation\footnote{For convenience,
the $(n-1)$-ary relation associated to $\forall v\in J\colon c$ is extended to an $n$-ary
relation by using the domain $J$ for the extra dimension.} associated to
the constraint $\forall v\in J\colon c$.

We first state a simple lemma, which will be used in proving the proposition to follow.

\begin{lemma}
  \label{lem:lemma-1}
  Given $n\in\NSet$, and two boxes $\Vector{B}=I_1\times\cdots\times I_n$ and 
  $\Vector{D}=J_1\times\cdots\times J_n$ such that $\Vector{B}\subseteq\Vector{D}$, the
  following does hold:
  \begin{equation*}
    \Vector{D}\setminus\Vector{B}=\bigcup_{k=1}^n J_1\times\cdots\times J_{k-1}\times 
    (J_k\setminus I_k)\times J_{k+1}\times \cdots\times J_n
  \end{equation*}
  In addition:
  \begin{equation*}
    \BoxesToSet{\BComplement{\Vector{D}}{\Vector{B}}}
    =\bigcup_{k=1}^n J_1\times\cdots\times J_{k-1}\times 
    \BoxesToSet{\BComplement{J_k}{I_k}}\times J_{k+1}\times \cdots\times J_n
  \end{equation*}

  \begin{proof}
    Immediate.
  \end{proof}
\end{lemma}

\begin{proposition}[Soundness of \algo{ICO2}]
  Let $c(x_1,\dots,x_{n-1},v)$ be an $n$-ary constraint, 
  $\Vector{B}=I_1\times\cdots\times I_{n-1}\times J$ a box, $\Gamma$ an outer contracting
operator for $c$, and 
$(\GSet{U}_i,\GSet{U}_o,\GSet{U}_u)=\algo{ICO2}_c^{\Gamma,(v,J)}(\Vector{B})$.
The following properties do hold:
\begin{equation}\label{eq:soundness:ICO2}
  \left\{\begin{array}{ll}
      \forall\Vector{D}\in\GSet{U}_o\colon&
      \Vector{D}\cap\CRel{c^\forall}=\emptyset\\
      \forall\Vector{D}\in\GSet{U}_i\colon& \Vector{D}\subseteq\CRel{c^\forall}
      \end{array}\right.
\end{equation}
\begin{proof}
  Termination of Alg.~\algo{ICO2} depends on a reasonable choice for
  the functions \algo{StoppingCriterion()} and
  $\algo{Split}^{\setminus v}_k\algo{()}$. More specifically, termination is
  ensured whenever $\algo{Split}^{\setminus v}_k\algo{()}$ creates boxes
  $\Vector{B_1},\dots,\Vector{B_k}$ that are all strictly smaller than
  \Vector{B''}, and when \algo{StoppingCriterion()} checks for the size of all
dimensions of \Vector{B''} but the one of $v$ being smaller than some threshold
that is attainable with the floating-point format at hand.

\medskip In order to prove the proposition, we have to show that the
properties given by Eq.~\eqref{eq:soundness:ICO2} always hold on lines
9 and 16 in Table~\ref{algo:ICO2}.  Provided that boxes put into
$\GSet{U}_o$ and $\GSet{U}_i$ on lines 5 and 7, respectively, verify
Eq.~\eqref{eq:soundness:ICO2}, it is not necessary to investigate what
is returned on line 13 since it is the mere union of the sets
previously computed.

\medskip
\noindent$\blacktriangleright$ We first consider the case when 
$\Domain{v}{\Vector{B'}} \neq \Domain{v}{\Vector{B}}$. Let
$\Vector{B}=I_1\times\cdots\times I_n$. By hypothesis,
$\Domain{v}{\Vector{B}}=I_n=J$ during the first call of \algo{ICO2}.
By completeness of $\Gamma_c$, if
$\Domain{v}{\Vector{B'}}\varsubsetneq\Domain{v}{\Vector{B}}$, there
exists a value in
$\Domain{v}{\Vector{B}}\setminus\Domain{v}{\Vector{B'}}$ such that $c$
does not hold when the free variables take their values in
$I_1\times\cdots\times I_{n-1}$. Consequently,
$\Vector{B}\cap\CRel{c^\forall}=\emptyset$.  The property does hold
for all the subsequent calls of \algo{ICO2} since the corresponding
domains for $v$ are always included in $J$ (by contractance of the
narrowing operators used).  Since
$\emptyset\subseteq\CRel{c^\forall}$, we have proved that the triplet
returned on Line~16 verifies Eq.~\eqref{eq:soundness:ICO2} for any
call of \algo{ICO2}.

\medskip
\noindent$\blacktriangleright$ We now consider the case when $\Domain{v}{\Vector{B'}} = \Domain{v}{\Vector{B}}$.

Let us prove that for any call $l$ of \algo{ICO2}, the assignments
performed on lines 5 and 7 are such that:
\begin{equation*}
  \left\{\begin{array}{l}
      \BoxesToSet{\GSet{U}_o^{(l)}}\cap\CRel{c^\forall}=\emptyset\\
      \BoxesToSet{\GSet{U}_i^{(l)}}\subseteq\CRel{c^\forall}
    \end{array}
    \right.
\end{equation*}
This suffices to prove the proposition since what is returned for the case 
$\Domain{v}{\Vector{B'}} = \Domain{v}{\Vector{B}}$ is either $\GSet{U}_o^{(l)}$
and $\GSet{U}_i^{(l)}$, or the union, componentwise, of $\GSet{U}_o^{(l)}$ and $\GSet{U}_i^{(l)}$
with sets of boxes $\GSet{U}_o^{(m)}$ and $\GSet{U}_i^{(m)}$ ($m>l$) computed in subsequent 
calls, and since we have already proved that what is returned whenever 
$\Domain{v}{\Vector{B'}} \neq \Domain{v}{\Vector{B}}$ verifies the property.

\smallskip
\noindent$\rhd$ We first consider Line~5: by completeness of $\Gamma_c$, 
$\BoxesToSet{\BComplement{\Vector{B}}{\Vector{B'}}}\subseteq\CRel{\negation{c}}$. Hence,
$\BoxesToSet{\GSet{U}_o}\subseteq\CRel{\negation{c}}$, and then 
$\BoxesToSet{\GSet{U}_o}\cap\CRel{c^\forall}=\emptyset$.

\smallskip
\noindent$\rhd$ We now consider Line~7: we will first prove that for any call $l$ of \algo{ICO2}:
\begin{equation}
  \label{eq:p1}
  \BoxesToSet{%
    \BComplement{\replaceDom{\Vector{B}^{(l)}}
                            {\Domain{v}{\Vector{B}^{(l)}}}
                            {J}}{%
                            \Vector{B}^{(l)}}}\subseteq\CRel{c}
\end{equation}

\noindent$\bullet$ This is trivially true for the first call ($l=1$) since, by hypothesis, 
$\Domain{v}{\Vector{B}^{(1)}}=J$.

\noindent$\bullet$ Assume that Eq.~\eqref{eq:p1} does hold for the $l$-th call of \algo{ICO2}.
Since $\Vector{B'}\subseteq\Vector{B}$ and
$\Domain{v}{\Vector{B'}^{(l)}}=\Domain{v}{\Vector{B}^{(l)}}$, we have:
  \begin{equation}
    \label{eq:p2}
  \BoxesToSet{%
    \BComplement{\replaceDom{\Vector{B'}^{(l)}}
                            {\Domain{v}{\Vector{B'}^{(l)}}}
                            {J}}{%
                            \Vector{B'}^{(l)}}}\subseteq\CRel{c}
  \end{equation}
By completeness of \CRel{\negation{c}}, we have:
\begin{equation}
  \label{eq:p3}
  \BoxesToSet{%
    \BComplement{\replaceDom{\Vector{B''}^{(l)}}
                            {\Domain{v}{\Vector{B''}^{(l)}}}
                            {\Domain{v}{\Vector{B'}^{(l)}}}}{%
                            \Vector{B''}^{(l)}}}\subseteq\CRel{c}
\end{equation}
By contractance of $\Gamma_{\negation{c}}$, $\Vector{B''}^{(l)}\subseteq\Vector{B'}^{(l)}$.
From equations \eqref{eq:p3} and \eqref{eq:p2}, we deduce:
\begin{equation}
  \label{eq:p4}
  \BoxesToSet{%
    \BComplement{\replaceDom{\Vector{B''}^{(l)}}
                            {\Domain{v}{\Vector{B''}^{(l)}}}
                            {J}}{%
                            \Vector{B''}^{(l)}}}\subseteq\CRel{c}
\end{equation}
The box $\Vector{B''}^{(l)}$ is split on Line~11 in $k$ boxes
$\Vector{B_i}^{(l)}$ ($i\in\{1,\dots,k\}$), with
$\Domain{v}{\Vector{B_i}^{(l)}}=\Domain{v}{\Vector{B''}^{(l)}}$ (no
splitting on the domain of $v$).

Consequently, we obtain from Eq.~\eqref{eq:p4}:
\begin{equation*}
  \forall i\in\{1,\dots,k\}\colon\quad  
  \BoxesToSet{%
    \BComplement{\replaceDom{\Vector{B_i}^{(l)}}
                            {\Domain{v}{\Vector{B_i}^{(l)}}}
                            {J}}{%
                            \Vector{B_i}^{(l)}}}\subseteq\CRel{c}
\end{equation*}
From Line~12, it follows that Eq.~\eqref{eq:p1} does hold for $l+1,\dots,l+k$ whenever it does
hold for $l$.

\smallskip
Having proved that Eq.~\eqref{eq:p1} does hold for any call $l$, we
reconsider Line~7: by completeness of $\Gamma_{\negation{c}}$, it comes:
\begin{equation*}
  \BoxesToSet{\BComplement{\Vector{B'}}{\Vector{B''}}}\subseteq\CRel{c}
\end{equation*}
Hence:
\begin{equation}
  \label{eq:p5}
  \BoxesToSet{\BComplement{\Vector{B'}}{%
                \replaceDom{\Vector{B''}}%
                           {\Domain{v}{\Vector{B''}}}
                           {J}}}\subseteq\CRel{c}
\end{equation}
since $\replaceDom{\Vector{B''}}{\Domain{v}{\Vector{B''}}}{J}\supseteq\Vector{B''}$.

From Eq.~\eqref{eq:p2} and \eqref{eq:p5}, we deduce:
\begin{equation}
  \label{eq:p6}
  \BoxesToSet{\BComplement{\replaceDom{\Vector{B'}}%
                                      {\Domain{v}{\Vector{B'}}}%
                                      {J}}{%
                \replaceDom{\Vector{B''}}%
                           {\Domain{v}{\Vector{B''}}}
                           {J}}}\subseteq\CRel{c}
\end{equation}
We also know from Lemma~\ref{lem:lemma-1} that all the boxes in 
$\BComplement{\replaceDom{\Vector{B'}}%
                                      {\Domain{v}{\Vector{B'}}}%
                                      {J}}{%
                \replaceDom{\Vector{B''}}%
                           {\Domain{v}{\Vector{B''}}}
                           {J}}$ have $J$
as their last dimension. From this and Eq.~\eqref{eq:p6}, we obtain:
\begin{equation*}
  \BoxesToSet{\BComplement{\replaceDom{\Vector{B'}}%
                                      {\Domain{v}{\Vector{B'}}}%
                                      {J}}{%
                \replaceDom{\Vector{B''}}%
                           {\Domain{v}{\Vector{B''}}}
                           {J}}}\subseteq\CRel{c^\forall}
\end{equation*}
That is, $\BoxesToSet{\GSet{U}_i}\subseteq\CRel{c^\forall}$ for any call of \algo{ICO2}.
\end{proof}
\end{proposition}

Handling the Unidimensional Guaranteed Tuning Problem described
by Eq.~\eqref{eq:unidimensional-GTP} is done by the propagation
algorithm \algo{IPA} presented in Table~\ref{algo:IPA} as follows:
each constraint of the system is considered in turn together with the
sets of elements verifying all the constraints already considered; the
main point concerning \algo{IPA} lies in that \emph{each constraint
  needs only be taken into account once}, since after having been
considered for the first time, the elements remaining in the variable
domains are all solutions of the constraint. As a consequence,
narrowing some domain later does not require additional work.
Alg.~\algo{IPA} is parameterized by an inner contracting operator
$\Upsilon$. Solving the UGTP can be done by instantiating \algo{IPA}
with \algo{ICO2}. Using \algo{ICO1} instead leads to an algorithm
computing the inner approximation of the relation associated to the
conjunction of constraints in \GSet{S}.

\begin{acmtable}{\textwidth}
\medskip
\begin{tabbing}
123\=123\=123\=123\=123\=\kill
\NUM{1} \>$\algo{IPA}^\Upsilon\bigl(\kwd{in}\colon
            \text{a set of constraints }\GSet{S}, \Vector{B}\in\ISet^n$;
        $\kwd{out}\colon (\GSet{U}_i,\GSet{U}_o,\GSet{U}_u)\in\powerset{\ISet^n}^3\bigr)$\\
\NUM{2}\> \BEGIN\\
\NUM{3}\> \> $\GSet{T}\gets\{\Vector{B}\}$\\
\NUM{4}\> \> $(\GSet{U}_o,\GSet{U}_u)\gets(\emptyset,\emptyset)$\\
\NUM{5}\> \> \FOREACH{$c\in\GSet{S}$}\\
\NUM{6}\> \>\> $\GSet{U}_i\gets\emptyset$\\
\NUM{7}\> \>\> \FOREACH{$\Vector{D}\in\GSet{T}$}\\
\NUM{8}\> \>\>\> $(\GSet{U}_i,\GSet{U}_o,\GSet{U}_u)\gets(\GSet{U}_i,\GSet{U}_o,\GSet{U}_u)\uplus\Upsilon_c(\Vector{D})$\\
\NUM{9}\> \>\> \ENDFOREACH\\
\NUM{10}\> \>\> $\GSet{T}\gets\GSet{U}_i$\\
\NUM{11}\> \> \ENDFOREACH\\
\NUM{12}\> \> \RETURN{$(\GSet{U}_i,\GSet{U}_o,\GSet{U}_u)$}\\
\NUM{12}\> \END
\end{tabbing}
\caption{Inner propagation algorithm for a set of constraints \GSet{S} 
  based on the inner contracting operator $\Upsilon$}
\label{algo:IPA}
\end{acmtable}

\begin{proposition}[Soundness of Alg.~\algo{IPA}]
  Given a set of constraints $\GSet{S}=\{c_1,\dots,c_m\}$, a box
  $\Vector{B}\in\ISet^n$, an inner contracting operator $\Upsilon$, a
  variable $v$, an interval $J$, a set of outer contracting operators
  $\{\Gamma_1,\dots,\Gamma_m\}$ for, respectively, $c_1$, \dots,
  $c_m$, and
  $(\GSet{U}_i,\GSet{U}_o,\GSet{U}_u)=\algo{IPA}^\Upsilon(\GSet{S},\Vector{B})$,
  let us note \CRel{\GSet{S}}, the relation associated to the
    constraint $c_1\wedge\cdots\wedge c_m$ and
    \CRel{\GSet{S}^\forall} the relation associated to the constraint
    $\forall v\in J\colon c_1\wedge\cdots\wedge c_m$. We have:
\begin{equation*}
  \begin{array}{ll}
      \text{If}\:\: \Upsilon_{c_i}=\algo{ICO1}_{c_i}^{\Gamma_i}\colon&
      \left\{\begin{array}{ll}
          \forall\Vector{D}\in\GSet{U}_o\colon&
          \Vector{D}\cap\CRel{\GSet{S}}=\emptyset\\
          \forall\Vector{D}\in\GSet{U}_i\colon& 
          \Vector{D}\subseteq\CRel{\GSet{S}}
          \end{array}\right.\\[10pt]
      \text{If}\:\: \Upsilon_{c_i}=\algo{ICO2}_{c_i}^{\Gamma_i,(v,J)}\colon&
      \left\{\begin{array}{ll}
          \forall\Vector{D}\in\GSet{U}_o\colon&
          \Vector{D}\cap\CRel{\GSet{S}^\forall}=\emptyset\\
          \forall\Vector{D}\in\GSet{U}_i\colon& 
          \Vector{D}\subseteq\CRel{\GSet{S}^\forall}
          \end{array}\right.
      \end{array}
\end{equation*}
\begin{proof}
  For each constraint $c$, we consider only the boxes that were put
  in $\GSet{U}_i$ by the preceding constraint (Lines $7$ and $10$). By
  soundness of \algo{ICO1} (resp.\ \algo{ICO2}), when considering
  constraint $c_j$, $\GSet{U}_i$ then contains on line $10$ only the
  boxes \Vector{D} verifying
  $\Vector{D}\subseteq\CRel{c_1}\wedge\cdots\wedge\Vector{D}\subseteq\CRel{c_j}$
  (resp.\ 
  $\Vector{D}\subseteq\CRel{c_1^\forall}\wedge\cdots\wedge\Vector{D}\subseteq\CRel{c_j^\forall}$).

  Once again, by soundness of \algo{ICO1} (resp.\ \algo{ICO2}), when considering a constraint 
  $c_j$ on Line $5$, $\GSet{U}_o$ contains only boxes \Vector{D} such that there was a 
constraint $c_k$ with $k<j$ for which $\Vector{D}\cap\CRel{c_k}=\emptyset$ (resp.\
 $\Vector{D}\cap\CRel{c_k^\forall}=\emptyset$). Consequently, 
$\Vector{D}\cap(\CRel{c_1}\cap\cdots\cap\CRel{c_k}\cap\cdots\wedge\CRel{c_j})=\emptyset$) (resp.\
$\Vector{D}\cap(\CRel{c_1^\forall}\cap\cdots\cap\CRel{c_k^\forall}\cap\cdots\wedge\CRel{c_j^\forall})=\emptyset$).
\end{proof}
\end{proposition}

Note that it is straightforward to modify \algo{IPA} in order to be
able to solve constraints of the form $\forall v\in J^{(1)}\colon
c_1\wedge\cdots\wedge\forall v\in J^{(m)}\colon c_m$ by replacing the
set of constraints \GSet{S} by a set of pairs $(c_i,J^{(i)})$ and by
initializing accordingly the box passed to $\Upsilon_c$ on
Line~$8$. Since we have not encountered applications requiring such
extension, we will not consider it any more in the
following.

\section{Camera Control and the \emph{Virtual Cameraman Problem}}
\label{sec:virtual-cameraman-problem}

Camera control is of interest to many fields, from computer graphics
(visualization in virtual environments~\cite{Blinn:CGA88}) to robotics
(sensor planning~\cite{Abrams-Allen:RR97}), and
cinematography~\cite{Davenport-et-al:CGA91}. Whatever the activity,
the objective is always to provide the user with an adequate view of
some points of interest in a scene for a predefined duration.  In
sensor planning~\cite{Tarabanis:TR90} for instance, one is interested
in positioning a camera over a robot in order to be able to monitor
its work whatever the position of its manipulating hand may be.  The
targeted applications are mostly real-time ones, which implies seeking
for only one solution through some optimization process.  The
modelling adopted is usually very close to the camera representation
(involving for instance, the direct control of the camera parameters
through the use of sliders), thereby impeding inexperienced users from
predicting the exact behaviour of the controlled device.

Yet, some
authors~\cite{Gleicher-Witkin:SIGGRAPH92,Drucker-et-al:SI3DG92}
from computer graphics and cinematography fields have worked on
determining camera parameters from given properties of a desired
scene.  Once more, most of these works are concerned with the
computation of only one solution by means of an optimization
criterion.

Among these works, one may single out the attempt by Gleicher, which
permits using some constraints including the position of a
three-dimensional point on the screen, and the orientation of two
points along with their distance on the projected image. Higher level
of control is obtained through the ability to bound a point within a
region of the image or to bound the size of an object. Time
derivatives of the camera parameters are computed in order to satisfy
user-defined controls. The method is devoted to maintaining
user-defined constraints while manipulating camera parameters. Camera
motion in an animated scene is not treated.

Another approach~\cite{Jardillier-Languenou:EG98}---inspired by
Snyder's seminal work~\cite{Snyder:SIGGRAPH92}---to the camera motion
computation problem relies on interval arithmetic to take into account
multiple constraints and screen space properties. Satisfying camera
movement parameters---with respect to some given scene
description---are obtained through constraint solving. The constraints
involved are handled with Alg.~\algo{JLA} (Tab.~\ref{algo:JLA}, 
p.~\pageref{algo:JLA}).

The main objective that motivated our work was to build a 
high-level tool allowing an artist to specify the desired camera
movements for a ``shot'' using \emph{cinematic primitives}. The
resulting description is then translated into a constraint system in
terms of the camera parameters, and solved using local consistency-based
pruning techniques. A huge set of solutions is usually output as a result,
from which a challenging task is to extract a limited representative
sample for presentation to the user.

The mathematical model used for representing the camera, its motion,
and the objects composing a scene is presented in the following section; then
follows the description of some of the cinematic primitives together with
their translation in terms of constraint systems. Addressing the problem of
the isolation of a representative solution sample is deferred until
Section~\ref{sec:improving-computation}.

\subsection{Modelling the Camera and the Objects}

A camera produces a 2D image by using a projection transformation of a
\emph{3D space scene}. In the following, the image is referred to as the
\emph{screen space} (or \emph{image space}) and the scene filmed as
the \emph{scene space} or \emph{3D space}.

The standard camera model is based on Euler angles to specify its
location and orientation. The work presented here is not bound to this
representation, though, and any other representation would be
convenient as well.

\begin{figure} 
\begin{center}  
  \includegraphics[width=.45\textwidth]{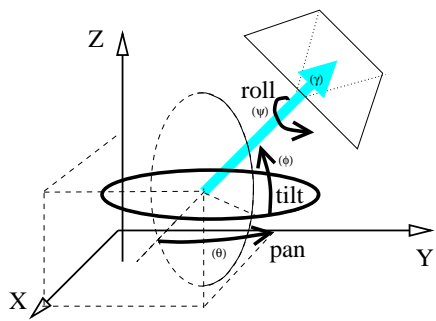}
  \caption{Camera model}
  \label{fig:camera-model} 
\end{center}
\end{figure}

A camera (Figure~\ref{fig:camera-model}) possesses seven degrees of
freedom, \textit{viz.}\ its Cartesian position in the space, its orientation,
and its focal length:

\begin{itemize}
\item \emph{Position.} Three scalars: $x$, $y$, and $z$;
\item \emph{View direction.} Three scalars: 
  \begin{itemize}
  \item \emph{Pan.} $\theta$,
  \item \emph{Tilt.} $\phi$,
  \item \emph{Roll.} $\psi$;
  \end{itemize}
\item \emph{Focal length.} One scalar: $\gamma$.
\end{itemize}

Most movies are made of a large number of short elementary ``shots''.
Therefore, a simple model for camera motion may be adopted without
losing too much expressiveness. Sophisticated camera movements are
obtained by assembling sequences of shots (Refer to Christie and al.'s
work~\cite{Christie-et-al:CP02} for a way to perform
this task).  Due to a lack of space, the following description of camera
movements (Figure~\ref{fig:camera-movements}) is restricted to
primitive ones, and the reader is referred to the excellent book by
Arijon~\cite{Arijon:76} for a thorough presentation of the others:

\begin{figure} 
  \begin{center} 
    \includegraphics[width=.9\textwidth]{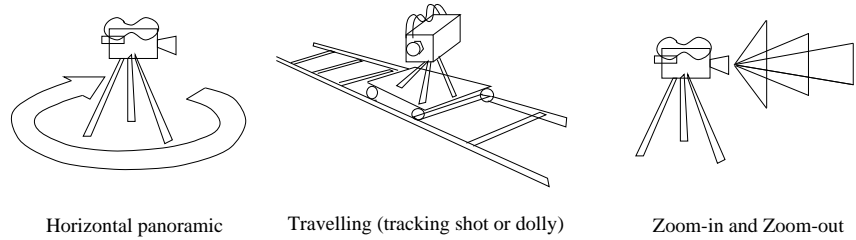}
    \caption{Camera movements}
    \label{fig:camera-movements} 
  \end{center}
\end{figure}

\begin{itemize}
\item\emph{Panoramic shot.} A panoramic shot may be horizontal or
  vertical (i.e.\ around vertical or horizontal axis). The camera
  location is usually constant;
\item\emph{Travelling (tracking shot or dolly).} A general term for a
  camera translation;
\item\emph{Zoom in or zoom out.} A variation of the focal
  length of the camera.
\end{itemize}

To our knowledge, existing declarative camera movement generators
compute camera animation frame by frame~\cite{Drucker:PhD94} or use
calculated key frames~\cite{Shoemake:CG85} (fixed camera location and
orientation), and interpolation of the camera parameters in-between.

In contrast, the tool described here is based on a
parametric representation. Computing a satisfying solution boils down
to determining all the camera parameters such that some constraints on
the scene are satisfied. Using the previous remark concerning simple
elementary shots, parameters are modelled with degree three
polynomials whose unknown is time $t$.  For example, an horizontal pan
(i.e.\ a movement along the horizontal orientation angle $\theta$)
might be defined as: $\theta(t) = c_{\theta}t + d_{\theta}$, where
$d_\theta$ represents the initial horizontal orientation of the camera
(at time $t=0$) and $c_\theta$ is a constant velocity.  Consequently,
given $V_{\theta}$ the maximum allowed pan velocity, one may note that
$c_{\theta}$ must lie in the domain \itv{-V_{\theta}}{V_{\theta}}, and
then $d_\theta$ must lie in the domain \itv{0}{2\pi}, in order to make
any orientation starting point eligible.

In our view, the scene is considered as a problem data. Hence, from
this point onward, every object composing a scene is assumed to have
its location, orientation and movement already set by the user.

Object properties rely on bounding volumes (location) and object axis
(orientation): each object is bounded by a volume called a
\emph{bounding box}. Compounds' bounding box is the smallest box
containing the bounding boxes of all the objects involved.  In
addition, bounding boxes may be associated to any set of objects (like
a group of characters). Many geometric modellers provide such a
hierarchy of bounding volumes.

\begin{figure} 
\begin{center}
  \includegraphics[width=.4\textwidth]{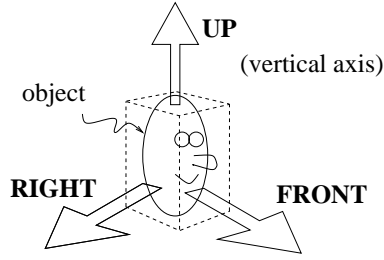}
\caption{Object axis and bounding volume}
\label{fig:object-axis} 
\end{center}
\end{figure}

The orientation of any object on the image is determined by three
vectors: \textbf{\sffamily Front}, \textbf{\sffamily Up} and
\textbf{\sffamily Right} (Figure~\ref{fig:object-axis}).

\subsection{Properties and Constraints}
\label{sec:properties}

The translation of declarative descriptions of scenes into constraints
is now presented. Three kinds of desired properties may be distinguished:
\begin{enumerate}
\item Properties on the camera;
\item Properties on the object screen location;
\item Properties on the object screen orientation.
\end{enumerate}

Camera properties are a means to set constraints on the camera motion.
Translation of properties on the objects of a scene is described
below.  The notion of \emph{frame} is first introduced as an aid to
constrain an object location on the screen: a \emph{frame} is a
rectangle whose borders are parallel to the screen borders. A frame
may be inside, outside, or partially inside the screen, though
it is usually fully contained in the screen.  For special purposes, the
frame size or/and location may be modified during the animation.

With frames, an artist can define the precise projection zone of an
object from the scene (3D) to the screen (2D).  Our belief in a
\emph{creation helping tool} leads us to prefer this kind of soft
constraint to the exact 2D screen location of the projection of a 3D
point.

\begin{figure}
  \begin{center}
    \hspace*{-1cm}\includegraphics[width=.7\textwidth]{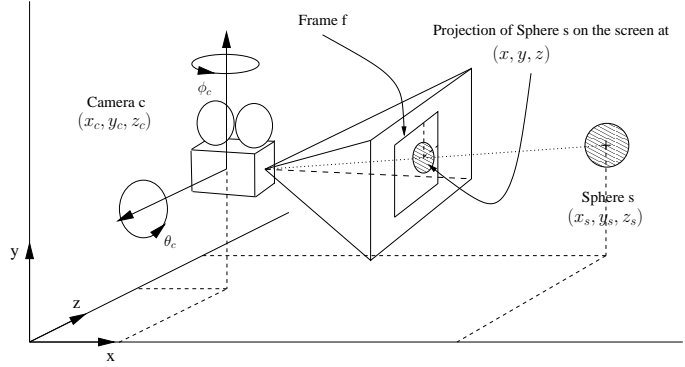}
    \caption{Projection in a frame}
    \label{fig:projection-frame}
  \end{center}
\end{figure}

The user defines frames (interactively or off-line), then chooses
properties to apply to a frame and an object. For example, the
statement (Figure~\ref{fig:projection-frame}):
\begin{quote}\itshape
  ``The sphere \object{s} with center
  $(x_\object{s}(t),y_\object{s}(t),z_\object{s}(t))$ and radius $r$
  must be fully included in the frame \object{f} defined by the
  bottom-left point $(x^1_\object{f}(t),y^1_\object{f}(t))$ and the
  top-right point $(x^2_\object{f}(t),y^2_\object{f}(t))$ during the
  20 seconds of the shot filmed by the camera \object{c} located at
  $(x_\object{c}(t),y_\object{c}(t),z_\object{c}(t))$ with orientation
  $\theta_\object{c}(t)$ and $\phi_\object{c}(t)$.''
\end{quote}
is translated into the nonlinear constraint system of equations and
inequations:
\begin{equation*}
\forall t\in\itv{0}{20}\colon
\left\{\begin{array}{ll}
  x^1_\object{f}(t) &\leq \bigl(x(t)+r\bigr)/\bigl(z(t)/\gamma_\object{c}(t)\bigr)\\
  x^2_\object{f}(t) &\geq \bigl(x(t)-r\bigr)/\bigl(z(t)/\gamma_\object{c}(t)\bigr)\\
  y^1_\object{f}(t) &\leq \bigl(y(t)+r\bigr)/\bigl(z(t)/\gamma_\object{c}(t)\bigr)\\
  y^2_\object{f}(t) &\geq \bigl(y(t)-r\bigr)/\bigl(z(t)/\gamma_\object{c}(t)\bigr) 
\end{array}\right.
\end{equation*}
with:
\begin{equation*}
\left\{\begin{array}{ll}
  x(t) &=-\bigl(x_\object{s}(t)-x_\object{c}(t)\bigr)\sin\theta_\object{c}(t)
  +\bigl(y_\object{s}(t)-y_\object{c}(t)\bigr)\cos\theta_\object{c}(t) \\
  y(t) &=-\bigl(x_\object{s}(t)-x_\object{c}(t)\bigr)\cos\theta_\object{c}(t)
  \sin\phi_\object{c}(t)+\\
  & \hspace{1cm}\bigl(y_\object{s}(t)-y_\object{c}(t)\bigr)
  \sin\phi_\object{c}(t)\sin\theta_\object{c}(t)
  +\bigl(z_\object{s}(t)-z_\object{c}(t)\bigr)\cos\phi_\object{c}(t) \\
  z(t) &=-\bigl(x_\object{s}(t)-x_\object{c}(t)\bigr)\cos\theta_\object{c}(t)
  \cos\phi_\object{c}(t)+\\
  & \hspace{1cm}\bigl(y_\object{s}(t)-y_\object{c}(t)\bigr)
  \sin\theta_\object{c}(t)\cos\phi_\object{c}(t)
  +\bigl(z_\object{s}(t)-z_\object{c}(t)\bigr)\sin\phi_\object{c}(t)
\end{array}\right.
\end{equation*}
where $\bigl(x(t),y(t),z(t)\bigr)$ is the projection of the center of \object{s}
in the screen space at time $t$.

Note that, in order to fall back to an instance of the guaranteed
tuning problem, it is possible to discard the three equations by
replacing $x(t)$, $y(t)$ and $z(t)$ in the inequalities by the
corresponding right-hand side.

The constraints resulting from the translation of the declarative
description of ``shots'' contain occurrences of the universally
quantified time variable $t$. 

\section{Examples and Benchmarking}
\label{sec:benchmarking}

\begin{figure}
  \begin{center}
  \includegraphics[width=\textwidth]{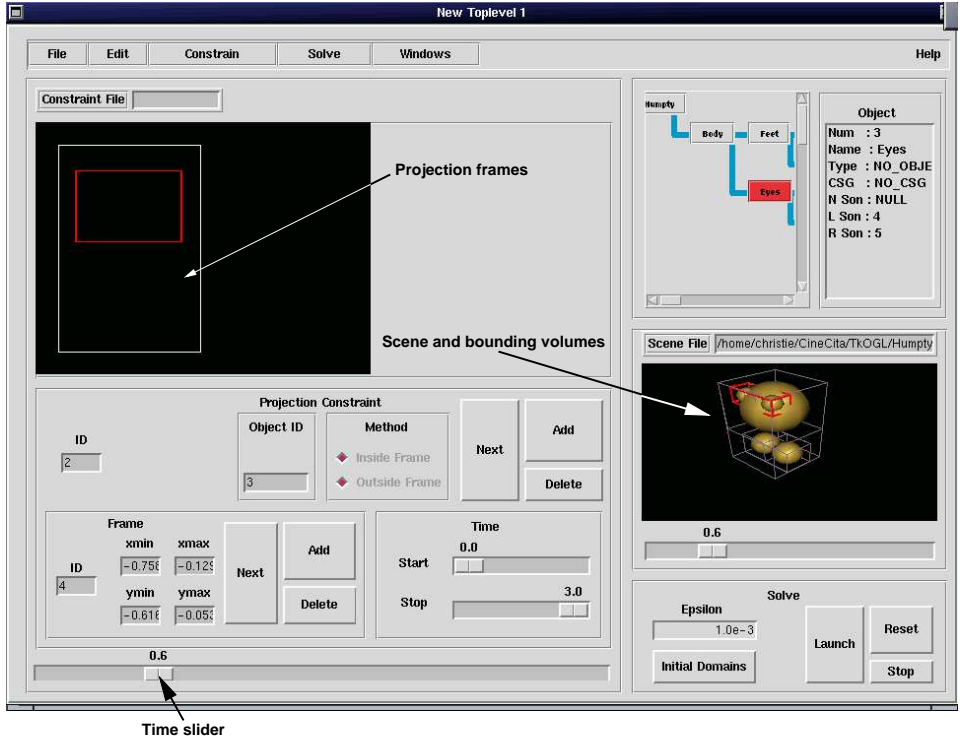}
  \caption{The declarative modeller tool}
  \label{fig:interface}
  \end{center}
\end{figure}

A high-level declarative modeller tool for camera motion has been
devised in order to validate the algorithms presented in
Section~\ref{sec:sound-constraint-solving}. The prototype is
written in \cplusplus\ and Tcl/Tk; Figure~\ref{fig:interface} presents
its graphical user interface: the animated scene to be filmed is
displayed in a window together with the bounding volumes, while
another window contains some previously drawn projection frames. The
user constructs frames, selects objects, and assigns properties to
objects in the scene.  The output is a set of satisfying camera paths,
and corresponding animations are shown in an output window.

In the next section, we present some problems used to assess the
quality of Alg.~$\algo{IPA}^\Upsilon$ (Tab.~\ref{algo:IPA}) using
Alg.~$\algo{ICO2}^{\algo{BC3r}}$ (Tab.~\ref{algo:ICO2} and 
Def.~\ref{def:outer-box-operator}) as the inner
contracting operator parameter $\Upsilon$ (hereafter referred as
\algo{IPABC}).  Comparisons with Alg.~\algo{JLA}
(Tab.~\ref{algo:JLA}) used by Jardillier and
Langu\'enou~\cite{Jardillier-Languenou:EG98} for the same kind of
applications are produced and commented. Some techniques to speed-up
the computation and improve the representativeness of the boxes output
are also described.

\subsection{Description of the benchmarks}
\label{sec:description-benchmarks}

\benchmark{\emph{Parabola}: a curve fitting problem}

This simple benchmark~\cite{Jardillier-Languenou:EG98} corresponds to
finding all the parabolas lying above a given line:
\begin{equation*}
  \forall t\in\itv{0}{2}\colon at^2+bt+c \geq 2t-1 \quad\text{with}\quad 
  a \in \itv{0}{1}, b \in \itv{0}{1}, c \in \itv{0}{1} 
\end{equation*}

\benchmark{\emph{Circle}: a trivial collision problem}

Benchmark \emph{Circle} is a collision problem: given $B$
a point moving along a circling path, find all points $A$ such that
the distance between $A$ and $B$ is always greater than a given value.
Benchmarks \emph{Circle$_2$} and \emph{Circle$_3$} are instances of the same problem with respectively
$2$ and $3$ points moving round in circles.  For only one circling
point, we have:

\begin{equation*}
  \begin{array}{l}
    \forall t\in\itv{-\pi}{\pi}\colon\\[2pt]
    \hspace{15pt}\sqrt{(r_1\sin t-x)^2+(r_1\cos t-y)^2} \geq d_1 
  \end{array}
  \quad\text{with}\quad
  \left\{\begin{array}{l}
      x \in \itv{-5}{5} \\
      y \in \itv{-5}{5} \\
      d_1 = 0.5 \\ 
    \end{array}\right.
  \end{equation*}

where $d_1$ is the mandatory minimal distance between $A$ and $B$, and
$r_1=2.5$ is the radius of $B$'s circling path.

\benchmark{\emph{Satellite}: a collision problem}

\begin{figure}
\begin{center}
\includegraphics[width=.9\textwidth]{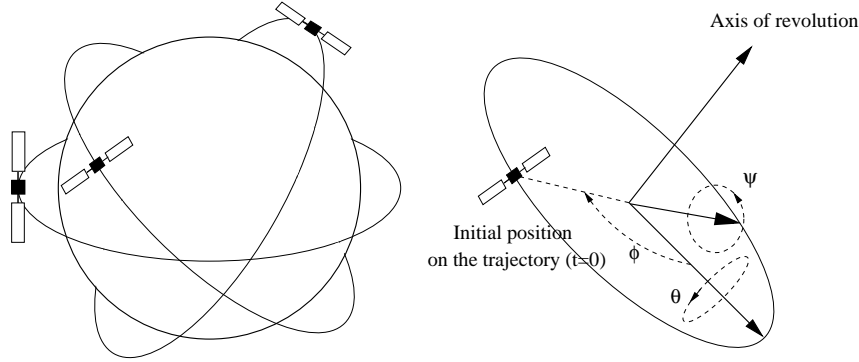}
\caption{A collision-free problem}
\label{fig:satellite}
\end{center}
\end{figure}

Given $n$ satellites swivelling around a planet
(Figure~\ref{fig:satellite}), we are looking for the parameters of an
$(n+1)$th trajectory on which to put another satellite. Obviously,
this trajectory must be such that the added satellite never collides
with the already launched ones.

The position of the $i$th satellite at time $t$ is defined by:
\begin{equation*}
\Vector{f_i}(t) =
\left(\begin{array}{l}
x_i(t) \\
y_i(t) \\
z_i(t) \\
\end{array}\right)
=
\left(\begin{array}{l}
d_i\cos{\theta_i}\sin{\omega_i t+\phi_i} \\
d_i\bigl(\sin{\psi_i}\sin{\theta_i}\sin{(\omega_i t+\phi_i)} +
\cos{\psi_i}\cos{(\omega_i t+\phi_i)}\bigr)\\
d_i\bigl(-\cos{\psi_i}\sin{\theta_i}\sin{(\omega_i t+\phi_i)} +
\sin{\psi_i}\cos{(\omega_i t+\phi_i)}\bigr)
\end{array}\right)
\end{equation*}
where variables $\theta_i$ and $\phi_i$ define the orientation of the plane of
revolution, $\omega_i$ the angular velocity of the satellite and $\phi_i$ its
initial position at $t=0$. Variable $d_i$ stands for the radius of the circle of
revolution. Considering $n$ satellites, we are looking for consistent values
of the unknowns $\theta_j,\phi_j,\omega_j,d_j$ of the satellite $j=n+1$ such
that:
\begin{equation*}
\forall t\in \itv{-\pi}{\pi}\colon
\left\{\begin{array}{l}
\dist{\Vector{f_1}(t)}{\Vector{f_j}(t)} \geq s \\
\dist{\Vector{f_2}(t)}{\Vector{f_j}(t)} \geq s \\
\vdots \\
\dist{\Vector{f_n}(t)}{\Vector{f_j}(t)} \geq s
\end{array}\right.
\end{equation*}
where $s$ represents the minimal distance allowed between two satellites.

The unknowns to be computed are $\theta_j,\phi_j$ and $\psi_j$, with
domains $\itv{0}{2\pi}$. In this benchmark, we consider three
satellites in the air with the following parameters:

\begin{center}
\medskip
\begin{tabular}{cccc}
\hline
Parameter & Satellite 1 & Satellite 2 & Satellite 3 \\
\hline
$d_i$ & 5.0 & 5.0 & 5.0 \\
$\omega_i$ & 1.0 & 1.0 & 1.0 \\
$\phi_i$ & 0.0 & 1.0 & 2.0 \\
$\theta_i$ & 0.0 & 1.0 & 1.5 \\
$\psi_i$ & 0.0 & 1.0 & 1.5
\end{tabular}
\medskip
\end{center}

\benchmark{\emph{Robot}: a collision problem}

This benchmark is based on Example~\ref{exa:robot},
p.~\pageref{exa:robot}. The actual constraint system to solve is:
\begin{equation*}
\forall t\in \itv{0}{2}\colon 
  \sqrt{\bigl(x - P_x(t)\bigr)^2 + \bigl(y - P_y(t)\bigr)^2} \geq d
\end{equation*}
where
\begin{equation*}
\left\{
\begin{array}{l}
P_x(t) = d_1\sin
\alpha_1(t)+d_2\sin\bigl(\alpha_1(t)+\alpha_2(t)-\pi\bigr)+
  d_3\sin\bigl(\alpha_1(t)+\alpha_2(t)+\alpha_3(t)\bigr)\\
P_y(t) = d_1\cos
\alpha_1(t)+d_2\cos\bigl(\alpha_1(t)+\alpha_2(t)-\pi\bigr)+
  d_3\cos\bigl(\alpha_1(t)+\alpha_2(t)+\alpha_3(t)\bigr)\\
\alpha_1(t) = t+\pi/4 \\
\alpha_2(t) = 2t - 1 \\
\alpha_3(t) = 0.2t+0.1
\end{array}\right.
\end{equation*}
The initial domains for the unknowns $x$ and $y$ are both set to \itv{0}{5},
the size of the robot's hand is set to $0.5$ and the respective
lengths of the arm's segments are $d_1=1.0$, $d_2=2.0$ and $d_3=1.0$.

\benchmark{\emph{PointPath}: a motion planning problem}

\begin{figure}
\begin{center}
\includegraphics[width=.7\textwidth]{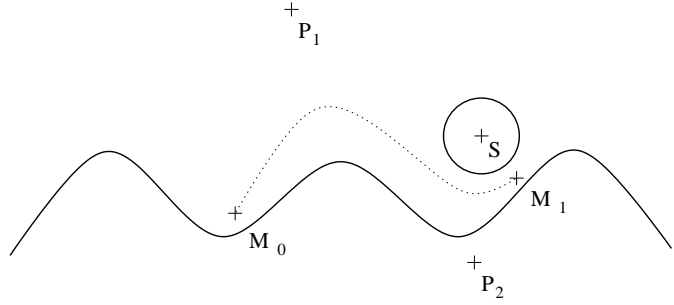}
\caption{Solution for the motion-planning problem}
\label{fig:PointPath}
\end{center}
\end{figure}

This benchmark~\cite{Jaulin-Walter:Automatica96} introduces a simple
motion planning problem (Figure~\ref{fig:PointPath}). We need to compute
an object's path, starting at position $M_0$ and ending at position
$M_1$, while avoiding collision with the ground and some objects. The
path $M(t)$ is tentatively chosen as a polynomial written as a linear
combination of Bernstein polynomials of degree 3, and controlled by
the points $P_1$ and $P_2$ of unknown coordinates:
\begin{equation*}
M(t) = M_0B^3_0(t) + P_1B^3_1(t) + P_2B^3_2(t) + M_1B^3_3(t)
\end{equation*}
where the Bernstein polynomials are:
\begin{equation*}
B^3_0(t) = (1-t)^3,\: B^3_1(t) = 3t(1-t)^2,\: B^3_2(t) = 3t^2(1-t),\: B^3_3 = t^3
\end{equation*}
The constraints specify that for all $t$ in the time frame:
\begin{enumerate}
\item $M(t)$ must be above a curve representing the floor;
\item the distance between $M(t)$ and a static object $S=(4.8,1.0)$ must be greater 
  than $1$.
\end{enumerate}
Which leads to:
\begin{equation*}
\forall t\in \itv{0}{1}\colon
\left\{
  \begin{array}{l}
    \bigl(x(t) - 4.8\bigr)^2 + \bigl(y(t) - 1\bigr)^2 \geq 1 \\
    y(t) \geq\sin\bigl(x(t)\bigr)\\    
  \end{array}\right.
\end{equation*}
with $M(t)=\displaystyle\binom{x(t)}{y(t)}$. Domains for the
coordinates of $P_1$ and $P_2$ are initialized to \itv{-10}{10}.

\benchmark{\emph{Projection}: a camera control problem}
Benchmark \emph{Projection} corresponds to the problem of projecting a moving sphere in
a moving frame, already presented in Section~\ref{sec:properties}. The initial values chosen 
for the camera are the following: $x_c\in\itv{-3}{3}$, $y_c\in\itv{-3}{3}$, 
$z_c=2$, $\phi_c\in\itv{-0.5}{0.5}$, $\theta_c=0$, and $\gamma_c=0.8$. 

\emph{Projection$_4$} is the original problem presented in
Section~\ref{sec:properties}; \emph{Projection$_5$} is the same
problem with one more constraint on the distance between the sphere
and the camera; \emph{Projection$_8$} is the problem with two frames
and two spheres.

\benchmark{\emph{GarloffGraf1}: inner approximation computation~\cite{Garloff-Graf:99}}

Find the values for the parameters $v$ and $w$  ensuring the stability of the polynomial:
\begin{equation*}
  p(s)=s^3+vs^2+(w-5v-13)s+w
\end{equation*}

Using the \emph{Linard-Chipart criterion}, the problem can be
transformed into the equivalent system:
\begin{align*}
  v,w&>0\\
  -5v^2-13v+vw-w&>0
\end{align*}
Following Garloff and Graf, we compute the inner approximation of the
relation $-5v^2-13v+vw-w>0$ for $v\in\itv{2}{10}$ and $w\in\itv{40}{50}$.

\benchmark{\emph{GarloffGraf2}: inner approximation computation~\cite{Garloff-Graf:99}}

This is again a stability problem: find sound values for $A$, $B$, and $D$ verifying:
{\small
\begin{align*}
  A,B,D&>0\\
  AB^2-D^2&>0\\
  -AB+A+D^2-D-1&>0\\
  AB-AD-2A+D^3+4D^2+4D&>0\\
  AB^3-AB^2D-4AB^2+2ABD+4AB+2BD^3+5BD^2+2BD-D^3-4D^2-4D&>0\\
  AB-2A-BD^2-4BD-4B-2D^2+3D-2&>0
\end{align*}}
Following Garloff and Graf, the initial domains chosen are
$A\in\itv{100}{120}$, $B\in\itv{0}{2}$, and $D\in\itv{10}{20}$. 

According to Abdallah and al.~\cite{Abdallah-et-al:MSCA96}, a
CAD-based software such as QEPCAD
(\url{http://www.cs.usna.edu/~qepcad/B/QEPCAD.html}) requires more
than two hours to prove the existence of a solution to the problem.

\subsection{Improving Computation}
\label{sec:improving-computation}

Solvers such as Numerica usually isolate solutions with variable
domains around $10^{-8}$ or $10^{-16}$ in width. By contrast, the
applications this paper focuses on are less demanding since the
resulting variable domains are used in the context of a display
screen, a ``low resolution'' device. In practice, one can consider
that a reasonable threshold $\varepsilon$ for the splitting process
during the search is some value lower or equal to $10^{-2}$ or $10^{-3}$.

\begin{figure}
\begin{center}
  \includegraphics[width=.9\hsize]{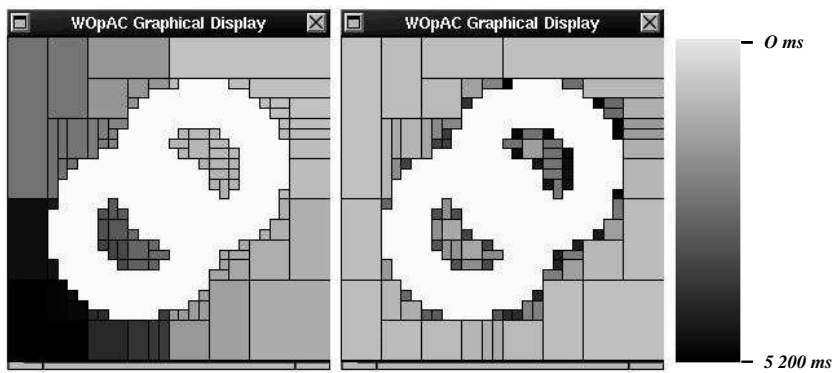}
 \caption{Depth-first vs.\ semi-depth-first}
\label{fig:nia-representativeness}
\end{center}
\end{figure}

One of the drawbacks of Algorithm
\algo{JLA}~\cite{Jardillier-Languenou:EG98} is that successive output
solutions are very similar, while it is of importance to be able to
provide the user with a representative sample of solutions as soon as
possible.  Tackling this problem using Alg.~\algo{IPABC} is done as
follows: given a constraint system of the form $\forall t\in I_t\colon
c_1\wedge\cdots\wedge c_m$ and a Cartesian product of domains
$\Vector{B}=I_1\times\cdots\times I_n$, we have two degrees of freedom
during the solving process, \emph{viz.}\ the selection of the next constraint
to consider, and the selection of the next variable to split.
Figure~\ref{fig:nia-representativeness} presents the differences with
regard to the order of generation of solutions for \emph{Circle$_2$}
for two strategies concerning the variable splitting order:
\begin{itemize}
\item\emph{depth-first}, where each constraint is considered in turn,
  and each domain is split to the threshold splitting limit;
\item \emph{semi-depth-first} where each constraint is considered in
  turn, but each variable is split only once and then put at the end of the
  domain queue.
\end{itemize}

As one may see, the semi-depth-first algorithm computes consecutive
solutions spread over all the search space, while the depth-first
algorithm computes solutions downward and from right to left.

Some strategies on constraint consideration order have also been
investigated, whose impact on speed is described hereunder. Four
strategies may be singled out:
\begin{description}
\item[Simple method] Box consistency is applied on the negation of
  each constraint in turn;
\item[Pre-parse method] This method interleaves the \emph{simple
    method} with a \emph{pre-parse algorithm}: given a constraint $c$,
  and $t$ the universally quantified variable, $k$ canonical intervals
  are extracted from the domain $I_t$ of $t$, and the consistency of $c$
  is tested for every one of them. At this stage, a failed check is
  sufficient to initiate a backtrack;
\item[Normal method] Box consistency is computed both for each
  constraint and for its negation;
\item[Global method] Given constraints $c_i,c_{i+1},\dots,c_m$, the
  \emph{global method} applies the \emph{normal method} on $c_i$, then
  checks whether the output boxes are consistent with the remaining
  constraints by mere evaluation. 
\end{description}

\begin{figure}
\begin{center}
\includegraphics[width=.44\textwidth]{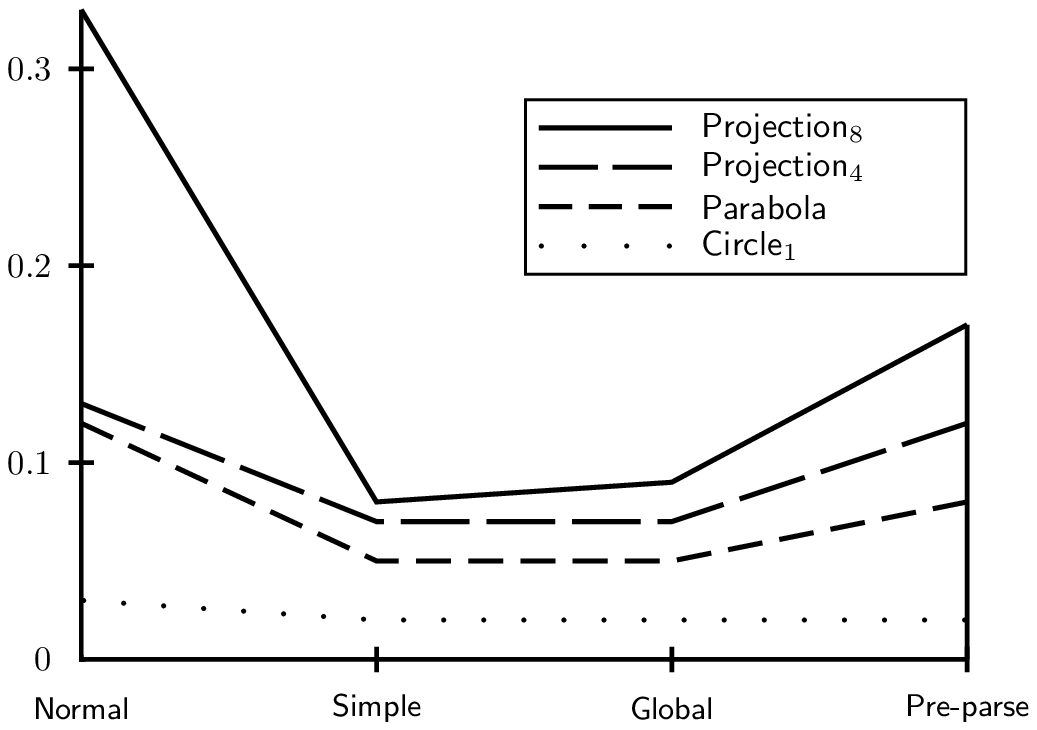}~\textbf{A.}
\includegraphics[width=.44\textwidth]{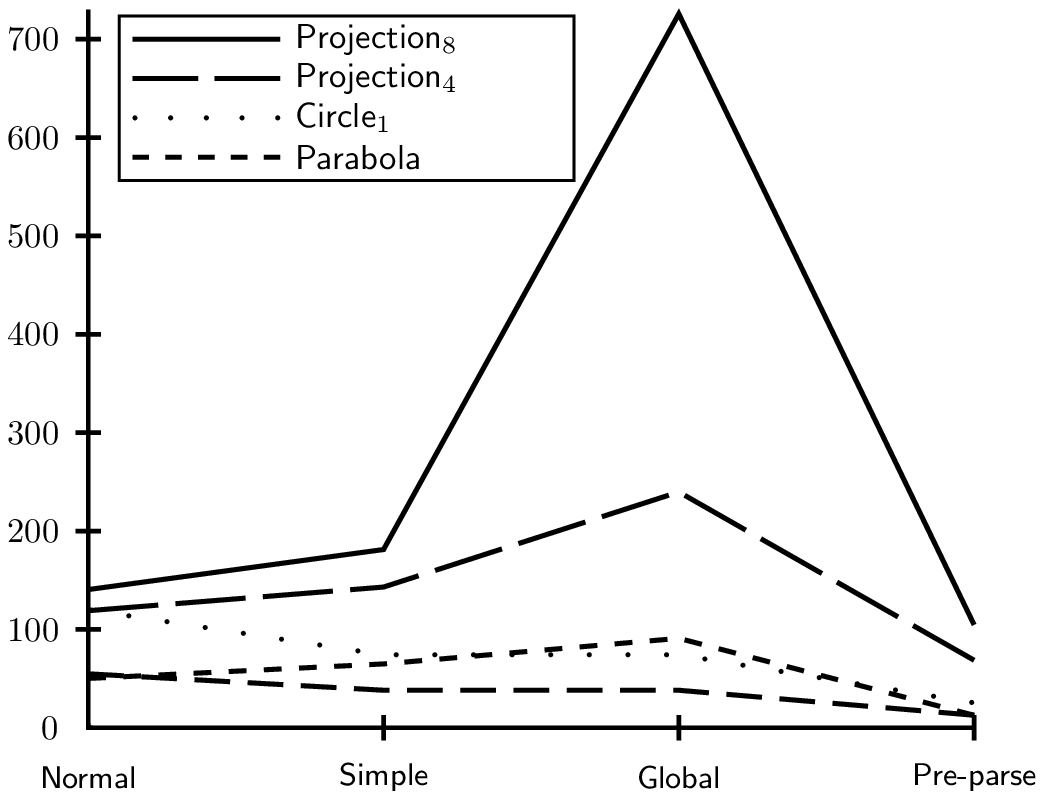}~\textbf{B.}
\caption{\textbf{A)} First solution. \textbf{B)} All solutions \emph{(times in seconds)}}
\label{fig:bench-1}
\end{center}
\end{figure}

Charts in Figure~\ref{fig:bench-1} present the time spent for obtaining the
first and all solutions for four benchmarks on a SUN UltraSparc 1/167\:MHz under
Solaris 2.5.

Considering first Chart~A, one may see that the \emph{simple method}
is the most interesting strategy for computing the first solution,
while the \emph{normal method} is very time consuming for problems
with ``many'' constraints (projection$_8$). On the other hand,
while the \emph{pre-parse method} is a bad choice for computing only
one solution, it is competitive for obtaining all solutions.

\subsection{Results}

\begin{figure}
  \begin{center}
    \includegraphics[width=.3\hsize]{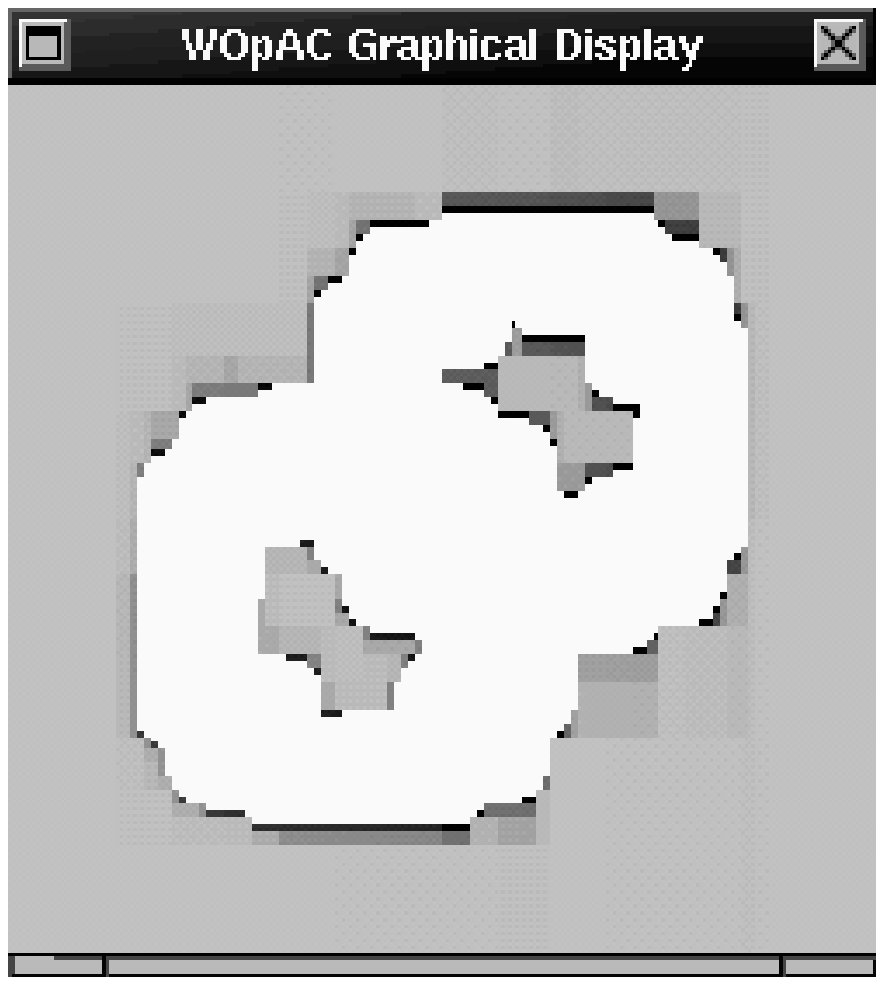}
    \vspace{0.5cm}
    \includegraphics[width=.3\hsize]{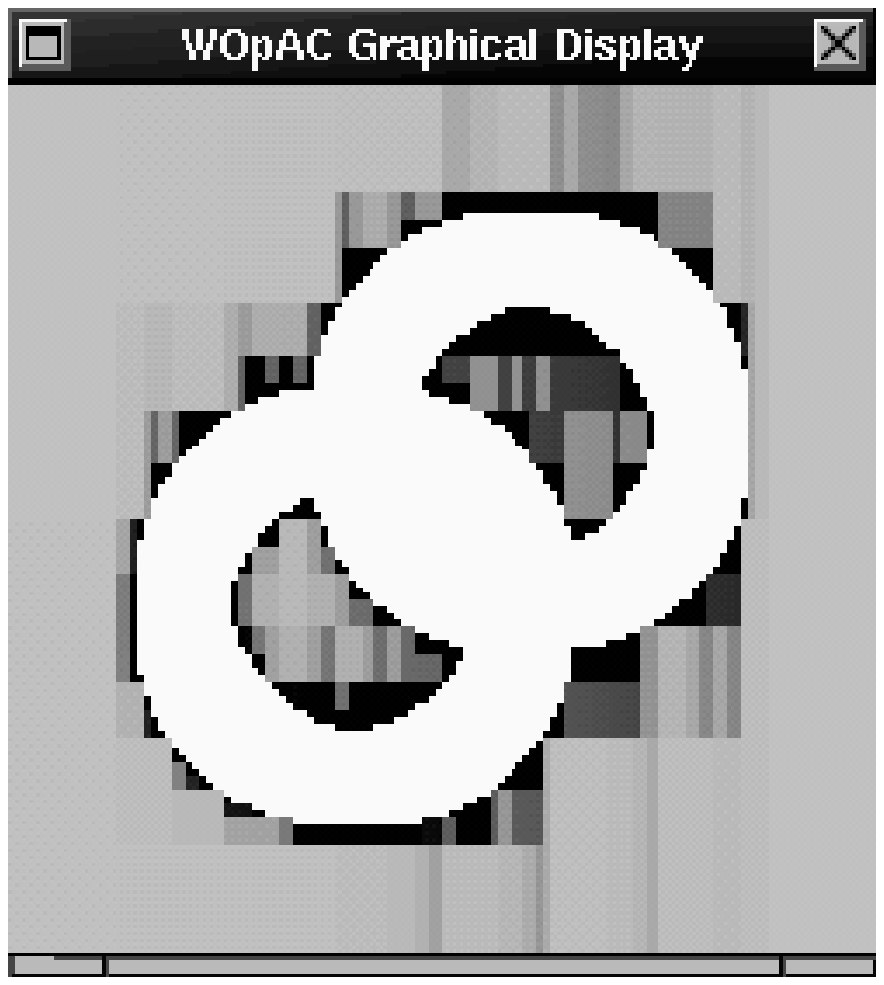}
    \vspace{0.5cm}
    \includegraphics[width=.3\hsize]{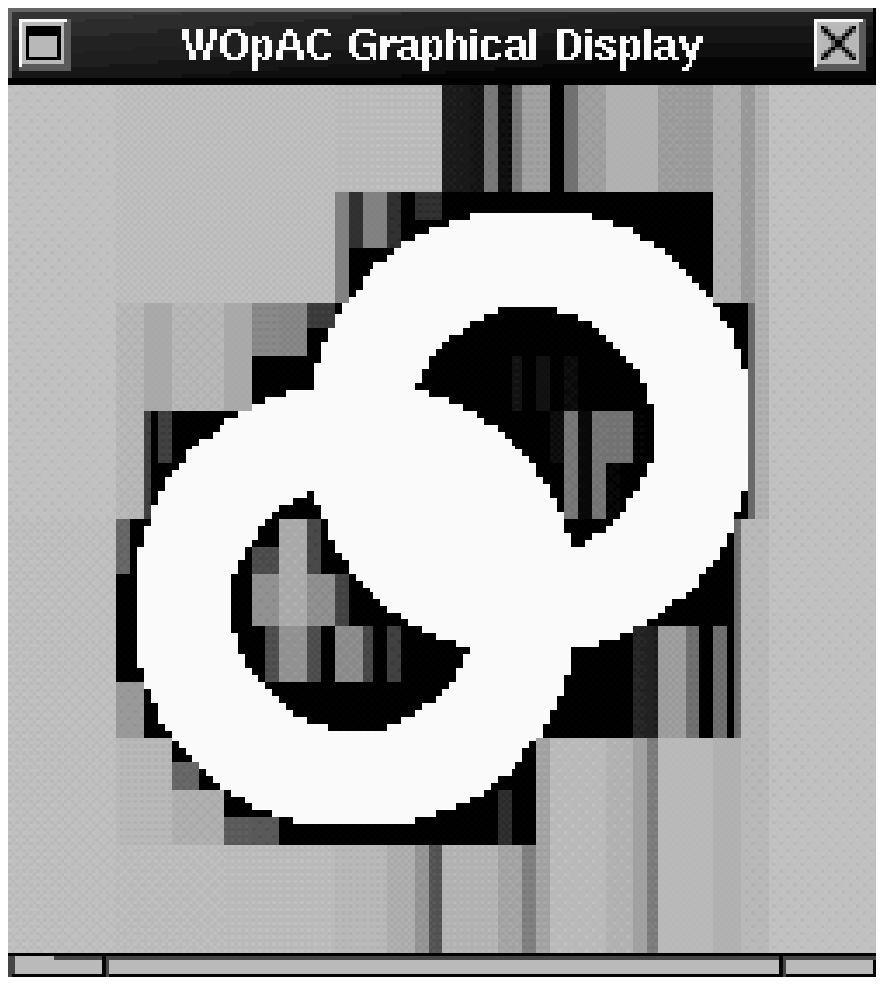}

    \caption{Impact of the time splitting threshold $\omega$ on precision in \algo{JLA} (from left to right: $\omega=0.5$, $\omega=0.1$, and $\omega=0.05$)}
    \label{fig:time-split}
  \end{center}
\end{figure}

Algorithms \algo{JLA} and \algo{IPABC} provide different sets of solutions for the same
problem. Consequently, a direct comparison of their performances is quite difficult. Moreover,
the actual implementation of \algo{JLA} uses a splitting threshold $\omega$ for slicing
the domain of the universally quantified variable $t$ instead of
checking consistency by eventually reaching canonicity of the samples
of the domain $I_t$. Figure~\ref{fig:time-split} shows the impact of the
threshold on the computed solutions for benchmark \emph{Circle$_2$}.

Table~\ref{tab:benchmark:01} (resp.\ \ref{tab:benchmark:02}) compares
algorithms \algo{JLA} and \algo{IPABC} from the speed point of
view for computing the first solution (resp.\ all solutions). Times
are given in seconds on a Pentium III at 800\,MHz under Linux.

\begin{acmtable}{\textwidth}
\medskip
\begin{center}
\begin{tabular}{lrrr}
\hline
Benchmark & \algo{JLA} & \algo{IPABC} & \algo{JLA}/\algo{IPABC}\\
\hline
Parabola $(\varepsilon=10^{-2})$ & 0.04 & 0.02 & 2.0\\
Parabola $(\varepsilon=10^{-3})$ & 0.77  & 0.02 & 38.5\\
Circle$_2$ $(\varepsilon=10^{-2})$ &  3.52 & 0.01 & 352000 \\
Circle$_2$ $(\varepsilon=10^{-3})$ &  3.57 & 0.01 & 357000 \\
Satellite $(\varepsilon=10^{-2})$ & 0.91 & 0.99 & 0.91 \\
Satellite $(\varepsilon=10^{-3})$ & 68.7 & 0.99 & 68.48 \\
Robot $(\varepsilon=10^{-1})$ & 0.13 & 0.01 & 13\\
Robot $(\varepsilon=10^{-2})$ & 0.50 & 0.01 & 50\\
Projection$_8$ $(\varepsilon=10^{-1})$ & 0.05 & 0.07 & 0.71 \\
Projection$_8$ $(\varepsilon=10^{-2})$ & 0.14 & 0.07 & 2.0 \\
Projection$_8$ $(\varepsilon=10^{-3})$ & 3.01 & 0.07 & 43.0\\
Projection$_4$ $(\varepsilon=10^{-2})$ & 0.11 & 0.03 & 3.66\\
Projection$_4$ $(\varepsilon=10^{-3})$ & 1.11 & 0.03 & 37\\
GarloffGraf1   $(\varepsilon=10^{-2})$ & 15.79  & 0.01 & 1579\\
GarloffGraf2   $(\varepsilon=10^{-2})$ & 5.58  & 0.04 & 139\\
PointPath      $(\varepsilon=0.5)$     & ???  & 7.85 & ???\\
\hline
\end{tabular}
\end{center}
\caption{\algo{JLA} vs.\ \algo{IPABC}---First solution} 
\label{tab:benchmark:01}
\end{acmtable}

As one can see, the efficiency of \algo{IPABC} compared to the one of \algo{JLA} increases
steadily with the precision required.

We were not able to obtain any result for \emph{PointPath} with
\algo{JLA} after several hours.

\begin{acmtable}{\textwidth}
\medskip
\begin{center}
\begin{tabular}{lrrr} 
\hline
Benchmark                &  \algo{JLA}  & \algo{IPABC} & \algo{JLA}/\algo{IPABC} \\
\hline
Parabola   & 10.65      & 1.22 & 8.72\\
Circle$_2$    & 3.16   & 0.15 & 21.06\\
Circle$_3$    & 3.41    & 0.55 & 6.2\\
Satellite   & $>$600 & 54.91  & $>$ 175 \\
Robot & 22.65 & 1.97 & 11.5\\
Projection$_8$ & $>$600.00 & 3.43 & $>$175.4\\
Projection$_5$ & 182.09 & 16.01 & 11.35\\
GarloffGraf1   $(\varepsilon=10^{-2})$ & 422.54  & 1.49 & 283.6\\
GarloffGraf2   $(\varepsilon=10^{-2})$ & 29.54  & 1.04 & 28.40\\
PointPath      $(\varepsilon=0.5)$     & ???  & 534.14 & ???\\
\hline
\end{tabular}
\end{center}
\caption{\algo{JLA} vs.\ \algo{IPABC}---All solutions}
\label{tab:benchmark:02}
\end{acmtable}

\section{Conclusion}
\label{sec:conclusion}

Unlike the methods used to deal with universally quantified variables
described in~\cite{Hong-Buchberger:TR91}, the algorithms presented in this paper
are purely numerical ones (except for the negation of constraints).
Since they rely on ``traditional'' techniques used by most of the
interval con\-straint-based solvers, they may benefit from the active
researches led to speed up these tools. What is more, they are 
applicable to the large range of constraints for which an
outer contracting operator may be devised. By contrast, CAD-based
methods deal with polynomial constraints only, as is the case with the
method based on Bernstein expansion~\cite{Garloff-Graf:99}. 

Despite the dramatic improvement of the new method described herein
over the one given by Jardillier and Langu\'enou, handling of complex
scenes with many objects and a camera allowed to move along all its
degrees of freedom in a reasonable time is beyond reach for the
moment. Nevertheless, a comforting idea is that most of the
traditional camera movements involve but few of the degrees of
freedom, thereby reducing the number of variables to consider.

Following the work of Markov~\cite{Markov:JUCS95} and
Shary~\cite{Shary:ISNMEB95}, a direction for future research is to
compare the use of \emph{Kaucher arithmetic}~\cite{Kaucher:CS80} to
compute inner approximations of relations with the use of
outer contracting operators. Their work is also related to the one by
Armengol et al.~\cite{Armengol-et-al:CCIA98} and Garde{\~{n}}es et
al.~\cite{Gardenes-Trepa:Computing80,Gardenes-Mielgo:86} on modal
interval arithmetic.  Yet, these approaches require algebraizing
trigonometric constraints, an operation known to slow down
computation~\cite{Pau-Schicho:JSC99}.

\appendix
\section*{APPENDIX}
\section{Notations}
{\footnotesize
\begin{tabular}{ll}
\FSet & Set of floating-point numbers\\
\nextFloat{a} & Smallest floating-point number greater than $a$\\
\prevFloat{a} & Greatest floating-point number smaller than $a$\\
\ISet & Set of closed intervals whose bounds are floating-point numbers\\
\Outer{\rho} & Smallest box containing $\rho$: $\Outer{\rho}=\bigcap\{\Vector{B}\in\ISet^n\mid\rho\subseteq\Vector{B}\}$\\
\Inner{\rho} & Inner approximation of $\rho$:\\
& $\Inner{\rho}=\{\Vector{r}\in\RSet^n \mid
  \Outer{\{\Vector{r}\}}\subseteq\rho\}$\\
\BoxesToSet{\GSet{S}} & Union of the boxes in \GSet{S}:\\
& $\BoxesToSet{\{\Vector{B_1},\dots,\Vector{B_n}\}}=\{\Vector{r}\in\RSet^n\mid
  \exists i\in\{1,\dots,n\}\text{ s.t. }\Vector{r}\in\Vector{B_i}\}$\\
\CRel{c} & Relation associated to the constraint $c$:\\
&$\CRel{c}=\{(r_1,\dots,r_n)\in\RSet^n\mid c(r_1,\dots,r_n)\}$\\
\CRel{c^\forall} & Relation associated to $\forall v\in I:c$\\
\negation{c} & ``Negation'' of the constraint $c$: ($\negation{f\leq0}$ is defined as
$f\geq0$)\\
\Domain{x}{\Vector{B}} & Domain of the variable $x$ in the box \Vector{B}\\
\BComplement{\Vector{B_1}}{\Vector{B_2}} & Set difference of \Vector{B_1} and
\Vector{B_2} as a set of boxes:\\
&$\BoxesToSet{\BComplement{\Vector{B_1}}{\Vector{B_2}}}=
  \{r\in\Vector{B_1}\mid\Outer{\{r\}}\cap\Vector{B_2}=\emptyset\}$\\
\replaceDom{\Vector{B}}{I}{J} & Box \Vector{B} where the interval $I$ is replaced by the 
interval $J$\\
\powerset{\GSet{S}} & Power set of the set \GSet{S}\\
$\algo{Split}_k(\Vector{B})$ & Split in $k$ boxes the box \Vector{B}\\
$\algo{Split}^{\setminus v}_k(\Vector{B})$ & Split in $k$ boxes the box \Vector{B} (never splits the interval corresponding to the \\
& domain of $v$\\
$\displaystyle\biguplus_i\{(\GSet{U}_1,\dots,\GSet{U}_n)\}$ & union of vectors of sets componentwise:\\
&$\displaystyle\biguplus_i\{(\GSet{U}_1^i,\dots,\GSet{U}_n^i)\}=(\displaystyle\bigcup_i\GSet{U}_1^i,\dots,
   \allowbreak\displaystyle\bigcup_i\GSet{U}_n^i)$\\
\end{tabular}
}

\begin{acks}
  The authors gratefully acknowledge the insightful perusal of the
  anonymous referees and discussions with Laurent Granvilliers and
  Lucas Bordeaux that helped improve preliminary versions of this
  paper.
\end{acks}


\begin{received}
Received July 2000;
revised September 2002;
accepted May 2003
\end{received}
\end{document}